\usepackage[american]{babel}
\usepackage[utf8]{inputenc} % allow utf-8 input
\usepackage[T1]{fontenc}    % use 8-bit T1 fonts
\usepackage{hyperref}       % hyperlinks
\usepackage{url}            % simple URL typesetting
\usepackage{booktabs}       % professional-quality tables
\usepackage{amsfonts}       % blackboard math symbols
\usepackage{nicefrac}       % compact symbols for 1/2, etc.
\usepackage{microtype}      % microtypography
\usepackage{xcolor}         % colors
\usepackage{pdfpages}       % plots
\usepackage{graphicx}       % plots
\usepackage{bm}             % bolds in math mode
\usepackage{tikz}           % check marks
\usepackage{algorithm}      % algorithms
\usepackage{amsmath}        % add math
\usepackage{algpseudocode}   % more algorithms
\usepackage{colortbl} 
\usepackage{amsmath,amssymb,amsthm,amsfonts,amsbsy,latexsym,dsfont,tikz}
\usepackage{booktabs}
\usepackage{float}
\usepackage[separate-uncertainty=true]{siunitx}
\usepackage{wrapfig,lipsum}
\usepackage{comment}
\usepackage{capt-of}% or \usepackage{caption}
\usepackage{varwidth}
\usepackage{natbib}
\usepackage{cuted}
\usepackage{dblfloatfix}
\usepackage{enumitem}

\newcolumntype{P}[1]{>{\centering\arraybackslash}p{#1}}

\theoremstyle{remark}
\newtheorem{thm}{Theorem}[section]
\newtheorem{lem}[thm]{Lemma}
\newtheorem{cor}[thm]{Corollary}

\newtheorem{defn}[thm]{Definition}

\renewcommand{\leq}{\leqslant} 
\renewcommand{\geq}{\geqslant}

\newcommand{\rd}{\mathrm{d}}

\renewcommand{\div}{\mbox{div}}
 
\newcommand{\cL}{\mathcal{L}}

\newcommand{\cP}{\mathcal{P}}

\newcommand{\vv}{\mathbf{v}}
 
\newcommand{\bR}{\mathbb{R}}

\DeclareMathOperator{\E}{\mathds{E}}

\DeclareMathOperator{\argmin}{argmin}

\definecolor{darkpastelgreen}{rgb}{0.01, 0.75, 0.24}
\DeclareRobustCommand{\greencheck}{%
  \tikz\fill[scale=0.4, color=darkpastelgreen]
  (0,.35) -- (.25,0) -- (1,.7) -- (.25,.15) -- cycle;%
}
\title{GeONet: a neural operator for learning the Wasserstein geodesic}

\author[1]{Andrew~Gracyk}
\author[2]{Xiaohui~Chen}
\affil[1]{%
     Department of Statistics\\
     University of Illinois at Urbana-Champaign\\
}
\affil[2]{%
    Department of Mathematics\\
    University of Southern California\\
}

\begin{document}

\maketitle

\begin{abstract}
   Optimal transport (OT) offers a versatile framework to compare complex data distributions in a geometrically meaningful way. Traditional methods for computing the Wasserstein distance and geodesic between probability measures require mesh-specific domain discretization and suffer from the curse-of-dimensionality. We present \emph{GeONet}, a mesh-invariant deep neural operator network that learns the non-linear mapping from the input pair of initial and terminal distributions to the Wasserstein geodesic connecting the two endpoint distributions. In the offline training stage, GeONet learns the saddle point optimality conditions for the dynamic formulation of the OT problem in the primal and dual spaces that are characterized by a coupled PDE system. The subsequent inference stage is instantaneous and can be deployed for real-time predictions in the online learning setting. We demonstrate that GeONet achieves comparable testing accuracy to the standard OT solvers on simulation examples and the MNIST dataset with considerably reduced inference-stage computational cost by orders of magnitude.
   \end{abstract}

  %In this paper, we propose a deep operator learning framework for Wasserstein geodesics. The proposed method is called optimal transport informed neural network (OTINN) that learns a neural operator that takes input of two probability measures and outputs a geodesic connecting them.
  
  %We present optimal transport informed neural networks (OTINNs), a neural network framework for learning Wasserstein geodesics that solve the Benamou-Brenier optimal transport formulation. The fabric of OTINNs stems from physics informed neural networks, where dual PINNs that take additional input parameters of initial and target distributions train simultaneously.

\section{Introduction}

Recent years have seen tremendous progress in statistical and computational optimal transport (OT) as a lens to explore machine learning problems. One prominent example is to use the Wasserstein distance to compare data distributions in a geometrically meaningful way, which has found various applications, such as in generative models~\citep{arjovsky17a}, domain adaptation~\citep{7586038} and computational geometry~\citep{Solomon_2015}. Computing the optimal transport map (if it exists) can be expressed in a fluid dynamics formulation with the minimum kinetic energy~\citep{Benamou2000ACF}. Such dynamical formulation defines geodesics in the Wasserstein space of probability measures, thus providing richer information for interpolating between data distributions that can be used to design efficient sampling methods from high-dimensional distributions~\citep{finlay2020train}. Moreover, the Wasserstein geodesic is also closely related to the optimal control theory~\citep{ChenGeorgiouPavon2021}, which has applications in robotics and control systems~\citep{8619816,9483194}.

Traditional methods for numerically computing the Wasserstein distance and geodesic require domain discretization that is often mesh-dependent (i.e., on regular grids or triangulated domains). Classical solvers such as Hungarian method~\citep{Kuhn1955}, the auction algorithm~\citep{BertsekaCastanon1989s}, and transportation simplex~\citep{LuenbergerYe2015}, suffer from the curse-of-dimensionality and scale poorly for even moderately mesh-sized problems~\citep{KlattTamelingMunk2020,Genevay2016,Benamou2000ACF}. Entropic regularized OT~\citep{NIPS2013_Cuturi} and the Sinkhorn algorithm~\citep{Sinkhorn_1964} have been shown to efficiently approximate the OT solutions at low computational cost, handling high-dimensional distributions~\citep{BenamouCarlierCuturiNennaPerye2015}; however, high accuracy is computationally obstructed with a small regularization parameter~\citep{Altschuler_2017,pmlr-v80-dvurechensky18a}. Recently, machine learning methods to compute the Wasserstein geodesic for a \emph{given} input pair of probability measures have been considered in~\citep{LiuMaChenZhaZhou2021,LiuGongLiu_2023,pooladian2023multisample,tong2023improving}, as well as \emph{amortized} methods~\cite{LacombeDigneCourtyBonneel_2023,amos2023meta} for generating static OT maps.

\begin{figure*}
  \centering
  \includegraphics[scale=0.64]{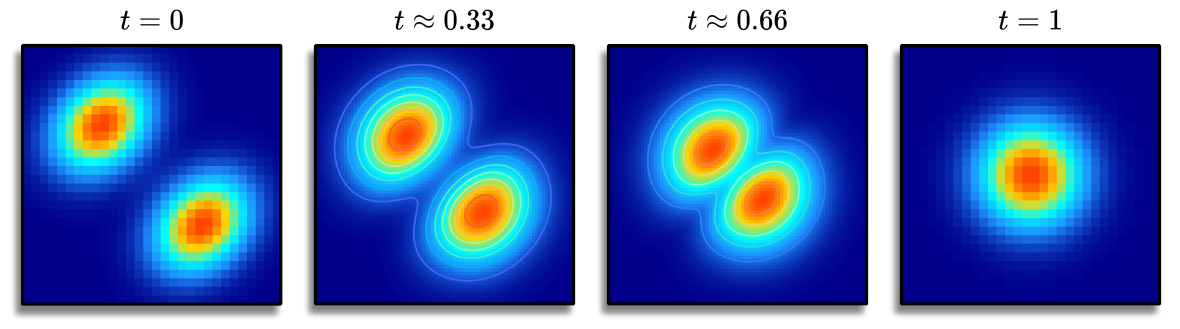}
  \vspace{-1mm}
  \caption{A geodesic at different spatial resolutions. Low-resolution inputs can be adapted into high-resolution geodesics (i.e., super-resolution) with our output mesh-invariant GeONet method.}
  \label{fig:GeONet_example}
  \vspace{-7mm}
\end{figure*}

A major challenge of using the OT-based techniques is that one needs to recompute the Wasserstein distance and geodesic for new input pair of probability measures. Thus, issues of scalability on large-scale datasets and suitability in the online learning setting are serious concerns for modern machine learning, computer graphics, and natural language processing tasks~\citep{Genevay2016,Solomon_2015,Kusner_2015ICML}. This motivates us to tackle the problem of learning the Wasserstein geodesic from an \emph{operator learning} perspective.

There is a recent line of work on learning neural operators for solving general differential equations or discovering equations from data, including DeepONet~\citep{LuJinPangZhangKarniadakis2021_DeepONet}, Fourier Neural Operators~\citep{Li_FourierNeuralOperator}, and physics-informed neural networks/operators (PINNs)~\citep{RAISSI2019686} and PINOs~\citep{https://doi.org/10.48550/arxiv.2111.03794}. Those methods are mesh-independent, data-driven, and designed to accommodate specific physical laws governed by certain partial differential equations (PDEs).

{\bf Our contributions.} In this paper, we propose a deep neural operator learning framework \emph{GeONet} for the Wasserstein geodesic. Our method is based on learning the optimality conditions in the dynamic formulation of the OT problem, which is characterized by a coupled PDE system in the primal and dual spaces. Our main idea is to recast the learning problem of the Wasserstein geodesic from training data into an operator learning problem for the solution of the PDEs corresponding to the primal and dual OT dynamics. Our method can learn the highly non-linear Wasserstein geodesic operator from a wide collection of training distributions. GeONet is also suitable for zero-shot super-resolution applications on images, i.e., it is trained on lower resolution and predicts at higher resolution without seeing any higher resolution data~\citep{ZSSR}. See Figure~\ref{fig:GeONet_example} for an example of the predicted higher-resolution Wasserstein geodesic connecting two lower-resolution Gaussian mixture distributions by GeONet.

Surprisingly, the training of our GeONet does not require the true geodesic data for connecting the two endpoint distributions. Instead, it only requires the training data as boundary pairs of initial and terminal distributions. The reason that GeONet needs much less input data is because its training process is implicitly informed by the OT dynamics such that the continuity equation in the primal space and Hamilton-Jacobi equation in the dual space must be simultaneously satisfied to ensure zero duality gap. Since the geodesic data are typically difficult to obtain without resorting to some traditional numerical solvers, the \emph{amortized inference} nature of GeONet, where inference on related training pairs can be reused~\citep{Gershman2014AmortizedII}, has substantial computational advantage over standard computational OT methods and machine learning methods for computing the geodesic designed for single input pair of distributions~\citep{COTFNT,LiuMaChenZhaZhou2021}. 

\begin{table*}
  \caption{We compare our method GeONet with other methodology, including traditional neural operators, physics-based neural networks for learning dynamics, and traditional optimal transport solvers. }
  
  %existing neural operators and networks for learning dynamics from data. PINN can be found in~\citep{RAISSI2019686} and the machine learning based minimax method can be found in~\citep{LiuMaChenZhaZhou2021}. We abbreviate neural operator as NO. We remark we do not check NO for satisfying the PDEs, but the true data is approximated, which do indeed satisfy the PDEs.}
  \label{tab:comparison}
  \centering
  \begin{tabular}[H]{
    p{4.6cm}  P{3.1cm} P{1.0cm}  P{1.8cm} P{1.2cm} }%{
  %  l
  %  S[table-format = 3]
  %  S[table-format = 2]
  %  S[table-format = 1.3]
  %  S[table-format = -2.2]
  %  S[table-format = 1.3]
  %  S[table-format = 1.3]
  %  S[table-format = 2.2]
  %  }
    \toprule  
    %\cmidrule(lr){1-5}   
    \textbf{Method characteristic} & {Neural operator w/o physics-informed learning} & {PINNs} & {Traditional OT solvers} & {GeONet (Ours)} \\
    \hline
    \text{operator learning} & {\greencheck}  &   &  & {\greencheck}  \\
    \text{satisfies the associated PDEs}   & {\greencheck} & {\greencheck} &
    {}  &  {\greencheck} \\
    \text{does not require known geodesic data} &  & {\greencheck} &  {\greencheck} &  {\greencheck} \\
    \text{output mesh independence} & {\greencheck}  & {\greencheck} &  {} & {\greencheck}  \\
    \bottomrule
\end{tabular}
\end{table*}

Once GeONet training is complete, the inference stage for predicting the geodesic connecting new initial and terminal data distributions requires only a forward pass of the network, and thus it can be performed in real time. In contrast, standard OT methods re-compute the Wasserstein distance and geodesic for each new input distribution pair. This is an appealing feature of amortized inference to use a pre-trained GeONet for fast geodesic computation or fine-tuning on a large number of future data distributions. Detailed comparison between our proposed method GeONet with other existing neural operators and networks for learning dynamics from data can be found in Table~\ref{tab:comparison}.

\begin{comment}
The rest of the paper is organized as follows. In Section~\ref{sec:background}, we review some background on static and dynamic formulations for the OT problem, as well as on the neural operator learning framework. In Section~\ref{sec:our_method}, we present the proposed GeONet method. In Section~\ref{sec:experiments}, we report the numeric performance on a synthetic experiment and a real image dataset.
\end{comment}

\section{Background}
\label{sec:background}

\subsection{Optimal transport problem: static and dynamic formulations}

The optimal mass transportation problem, first considered by the French engineer Gaspard Monge, is to find an optimal map $T^*$ for transporting a source distribution $\mu_0$ to a target distribution $\mu_1$ that minimizes some cost function $c : \bR^d \times \bR^d \to \bR$:
\begin{equation}
    \label{eqn:monge_problem}
    \min_{T : \bR^d \to \bR^d} \left\{ \int_{\bR^d} c(x, T(x)) \; \rd \mu_0(x) :  T_\sharp \mu_0 = \mu_1 \right\},
\end{equation}
where $T_\sharp \mu$ denotes the pushforward measure defined by $(T_\sharp \mu) (B) = \mu(T^{-1}(B))$ for measurable subset $B \subset \bR^d$. In this paper, we focus on the quadratic cost $c(x,y) = {1\over2} \|x-y\|_2^2$. The Monge problem~\eqref{eqn:monge_problem} induces a metric, known as the \emph{Wasserstein distance}, on the space $\cP_2(\bR^d)$ of probability measures on $\bR^d$ with finite second moments. In particular, the 2-Wasserstein distance can be expressed in the relaxed Kantorovich form:
\begin{equation}
    \label{eqn:kantorovich_problem}
    W_2^2(\mu_0, \mu_1) := \min_{\gamma \in \Gamma(\mu_0, \mu_1)} \left\{ \int_{\bR^d \times \bR^d} \|x - y\|_2^2 \; \rd \gamma(x,y) \right\},
\end{equation}
where minimization over $\gamma$ runs over all possible couplings $\Gamma(\mu_0, \mu_1)$ with marginal distributions $\mu_0$ and $\mu_1$. Problem~\eqref{eqn:kantorovich_problem} has the dual form (cf.~\cite{Villani2003_topics-in-ot})
\begin{align}
    \label{eqn:kantorovich_dual_problem}
    & {1\over 2} W_2^2(\mu_0, \mu_1)  = \sup_{\varphi \in L^1(\mu_0), \; \psi \in L^1(\mu_1)} \Big\{ \int_{\bR^d} \varphi \; \rd \mu_0 \\ & \ \ \ \ \ \ + \int_{\bR^d} \psi \; \rd \mu_1 : \varphi(x) + \psi(y) \leq {\|x-y\|_2^2 \over 2} \Big\}.
\end{align}
Problems~\eqref{eqn:monge_problem} and~\eqref{eqn:kantorovich_problem} are both referred as the \emph{static OT} problem, which has close connection to fluid dynamics. Specifically, the Benamou-Brenier dynamic formulation~\citep{Benamou2000ACF} expresses the Wasserstein distance as a minimal kinetic energy flow problem:
\begin{align}
\label{eqn:benamou-brenier_formula}
\begin{gathered}
{1\over2} W_2^2(\mu_0, \mu_1) = \min_{(\mu, \vv)} \int_0^1 \int_{\mathbb{R}^d} {1\over2} || \vv(x, t) ||_2^2 \ \mu(x, t) \ \rd x \ \rd t \\
  \mbox{subject to}  \ \  \partial_t \mu + \div(\mu \vv) = 0, \mu(\cdot, 0) = \mu_0, \mu(\cdot, 1) = \mu_1,
 \end{gathered}
\end{align}
where $\mu_t := \mu(\cdot, t)$ is the probability density flow at time $t$ satisfying the continuity equation (CE) constraint $\partial_t \mu + \div(\mu \vv) = 0$ that ensures the conservation of unit mass along the flow $(\mu_t)_{t \in [0,1]}$. To solve~\eqref{eqn:benamou-brenier_formula}, we apply the Lagrange multiplier method to find the saddle point in the primal and dual variables. In particular, for any flow $\mu_t$ initializing from $\mu_0$ and terminating at $\mu_1$, the Lagrangian function for~\eqref{eqn:benamou-brenier_formula} can be written as
\begin{equation}
    \label{eqn:lagrangian_benamou-brenier}
    \begin{gathered}
    \cL(\mu, \vv, u) = \int_0^1 \int_{\bR^d} \left[ {1\over2} \|\vv\|_2^2 \mu + \left( \partial_t \mu + \div(\mu \vv) \right) u \right] \; \rd x \; \rd t,
    \end{gathered}
\end{equation}
where $u := u(x, t)$ is the dual variable for the continuity equation. Using integration-by-parts under suitable decay condition for $\|x\|_2 \to \infty$, we find that the optimal dual variable $u^*$ satisfies the Hamilton-Jacobi (HJ) equation for the dynamic OT problem
\begin{equation}
    \label{eqn:hamilton-jacobi}
    \partial_t u + {1\over2} \|\nabla u\|_2^2 = 0,
\end{equation}
and the optimal velocity vector field is given by $\vv^*(x, t) = \nabla u^*(x, t)$. Hence, we obtained that the  Karush–Kuhn–Tucker (KKT) optimality conditions for~\eqref{eqn:lagrangian_benamou-brenier} are solution $(\mu^*, u^*)$ to the following system of PDEs:
\begin{equation}
    \label{eqn:benamou-brenier_kkt}
    \left\{
    \begin{gathered}
      \partial_t \mu + \div(\mu \nabla u) = 0, \ \ \partial_t u + {1\over2} \|\nabla u\|_2^2 = 0, \\
     \mu(\cdot,0) = \mu_0, \ \ \mu(\cdot,1) = \mu_1.
     \end{gathered} \right.
\end{equation}

In addition, solution to the Hamilton-Jacobi equation~\eqref{eqn:hamilton-jacobi} can be viewed as an interpolation $u(x,t)$ of the Kantorovich potential between the initial and terminal distributions in the sense that $u^*(x, 1) = \psi^*(x)$ and $u^*(x, 0) = -\varphi^*(x)$ (both up to some additive constants), where $\psi^*$ and $\varphi^*$ are the optimal Kantorovich potentials for solving the static dual OT problem~\eqref{eqn:kantorovich_dual_problem}. A detailed derivation of the primal-dual optimality conditions for the dynamical OT formulation is provided in Appendix~\ref{app:sec:primal-dual_optimality}.

\subsection{Learning neural operators}

A neural operator generalizes a neural network that learns a mapping $\Gamma^\dagger : {\cal A} \to {\cal U}$ between infinite-dimensional function spaces $\cal A$ and $\cal U$~\citep{Kovachki_neuraloperator,LiKovachkiAzizzadenesheliLiuBhattacharyaStuartAnandkumar}. Typically, $\cal A$ and $\cal U$ contain functions defined over a space-time domain $\Omega \times [0, T]$, where $\Omega$ is taken as a subset of $\bR^d$, and the mapping of interest $\Gamma^\dagger$ is implicitly defined through certain differential operator. For example, the physics informed neural network (PINN)~\citep{RAISSI2019686} aims to use a neural network to find a solution to the PDE 
\begin{equation}
    \label{eqn:pinn}
    \partial_t u + \mathcal{D}[u] = 0,
\end{equation}
given the boundary data $u(\cdot,0) = u_0$ and $u(\cdot,T) = u_T$, where $\mathcal{D} := {\cal D}(a)$ denotes a non-linear differential operator in space that may depend on the input function $a \in {\cal A}$. Different from the classical neural network learning paradigm that is purely data-driven, a PINN has less input data (i.e., some randomly sampled data points from the solution $u = \Gamma^\dagger(a)$ and the boundary conditions) since the solution operator $\Gamma^\dagger$ has to obeys the induced physical laws governed by~\eqref{eqn:pinn}. Even though the PINN is mesh-independent, it only learns the solution for a {\it single} instance of the input function $a$ in the PDE~\eqref{eqn:pinn}. In order to learn the dynamical behavior of the inverse problem $\Gamma^\dagger : {\cal A} \to {\cal U}$ for an entire family of $\cal A$, we consider the operator learning perspective.

The idea of using neural networks to approximate any non-linear continuous operator stems from the universal approximation theorem for operators~\citep{392253,LuJinPangZhangKarniadakis2021_DeepONet}. In particular, we construct a parametric map by a neural network $\Gamma : \mathcal{A}\times \Theta \rightarrow \mathcal{U}$ 
%\begin{equation}
%\label{eqn:Neural_operator}
%\Gamma : \mathcal{A}\times \Theta \rightarrow \mathcal{U}
%\end{equation}
for a finite-dimensional parameter space $\Theta$ to approximate the true solution operator $\Gamma^{\dag}$. In this paper, we adopt the \emph{DeepONet} architecture~\citep{LuJinPangZhangKarniadakis2021_DeepONet}, suitable for their ability to learn mappings from pairings of initial input data~\citep{https://doi.org/10.48550/arxiv.2202.08942}, to model $\Gamma$. We refer the readers to Appendix~\ref{app:sec:DeepONets} for some basics of DeepONet and its enhanced versions. Then, the neural operator learning problem is to find an optimal $\theta^* \in \Theta$ as a minimizer of the classical risk minimization problem
\begin{align}
\label{eqn:PINN_formulation_1}
\begin{gathered}
    \min_{\theta \in \Theta}  \E_{(a, u_0, u_T) \sim \mu} \Big[ \big\| (\partial_{t} + \mathcal{D} ) \Gamma(a, \theta ) \big\|_{L^2(\Omega \times (0,T)) }^2   \\  
  \qquad + \lambda_0 \big\|\Gamma(a,\theta)(\cdot,0) - u_0 \big\|_{L^2(\Omega)}^2 \\
  \qquad + \lambda_T \big\|\Gamma(a,\theta)(\cdot,T) - u_T \big\|_{L^2(\Omega)}^2  \Big],
\end{gathered}
\end{align}
where the input data $(a, u_0, u_T)$ are sampled from some joint distribution $\mu$. In~\eqref{eqn:PINN_formulation_1}, we minimize the PDE residual loss corresponding to $\partial_{t} u + \mathcal{D}[u] = 0$ while constraining the network by imposing boundary conditions. The loss function has weights $\lambda_0, \lambda_T > 0$. Given a finite sample $\{(a^{(i)}, u_0^{(i)}, u_T^{(i)})\}_{i=1}^n$, and data points randomly sampled in the space-time domain $\Omega \times (0, T)$, we may minimize the empirical loss analog of~\eqref{eqn:PINN_formulation_1} by replacing $\| \cdot \|_{L^2(\Omega \times (0,T))}$ with the discrete $L^2$ norm over domain $\Omega \times (0,T)$. Computation of the exact differential operators $\partial_t$ and $\cal D$ can be conveniently exploited via automatic differentiation in standard deep learning packages.

\section{Our method}
\label{sec:our_method}

%The associated PDEs and a PINN implementation provide the means to solve the Benamou-Brenier geodesic problem, from both of which does the design of our neural operator emerge.

\begin{figure*}[t]
  \centering
  \includegraphics[scale=0.54]{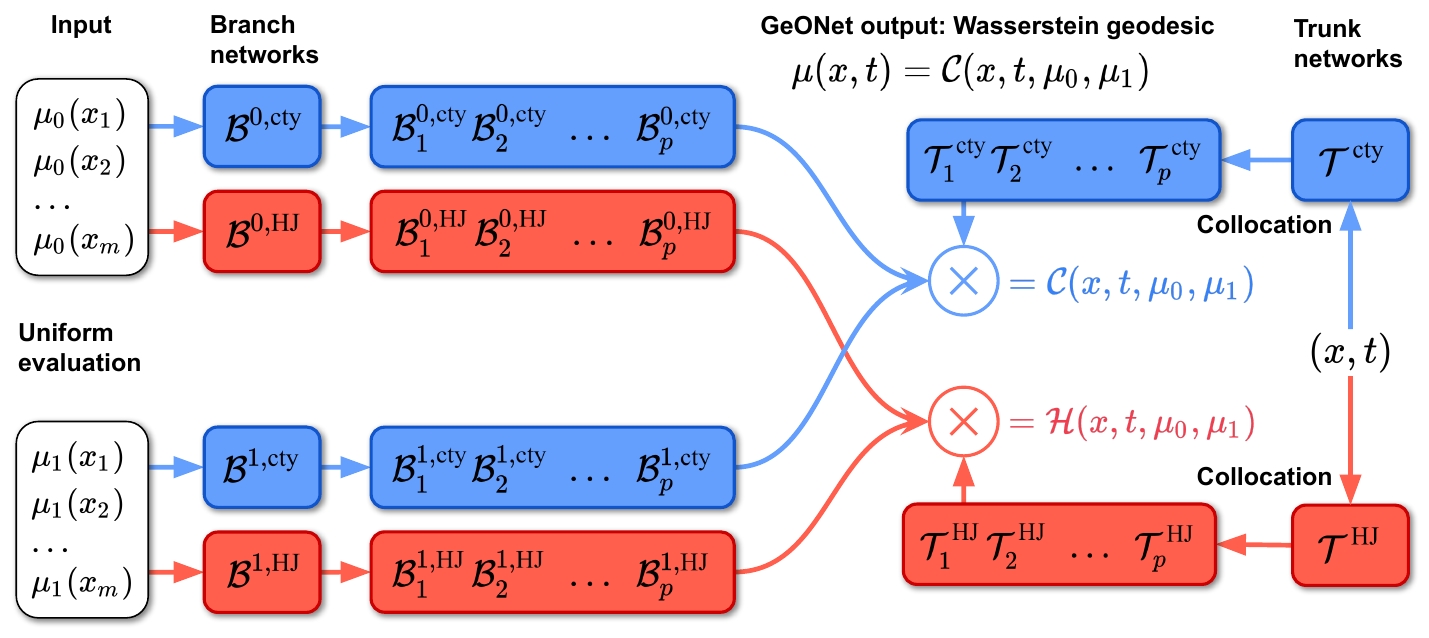}
  \caption{Architecture of GeONet, containing six neural networks to solve the continuity and Hamilton-Jacobi (HJ) equations, three for each. We minimize the total loss, and the continuity solution yields the geodesic. GeONet branches and trunks output vectors of dimension $p$, in which we perform multiplication among neural network elements  to produce the continuity and HJ solutions.}
  \label{fig:GeONet_diagram}
\end{figure*}

We present {\it GeONet}, a geodesic operator network for learning the Wasserstein geodesic $\{\mu_t\}_{t \in [0,1]}$ connecting $\mu_0$ to $\mu_1$ from the distance $W_2(\mu_0, \mu_1)$. Let $\Omega \subset \bR^d$ be the spatial domain where the probability measures are supported. For probability measures $\mu_0, \mu_1 \in {\cal P}_2(\Omega)$, it is well-known that the constant-speed geodesic $\{\mu_t\}_{t \in [0,1]}$ between $\mu_0$ and $\mu_1$ is an absolutely continuous curve in the metric space $({\cal P}_2(\Omega), W_2)$, which we denote as $\text{AC}({\cal P}_2(\Omega))$. $\mu_t$ solves the kinetic energy minimization problem in~\eqref{eqn:benamou-brenier_formula}~\citep{sabtanbrogio2015_OT}. Some basic facts on the metric geometry structure of the Wasserstein geodesic and its relation to the fluid dynamic formulation are reviewed and discussed in Appendix~\ref{app:sec:wasserstein_facts}. In this work, our goal is to learn a non-linear operator
\begin{align}
    \Gamma^\dagger : {\cal P}_{2}(\Omega) \times {\cal P}_{2}(\Omega) \to & \ \text{AC}({\cal P}_2(\Omega)), \\
    (\mu_0, \mu_1) \mapsto & \ \{\mu_t\}_{t \in [0,1]},
\end{align}
based on a training dataset $\{(\mu_0^{(1)}, \mu_1^{(1)}), \hdots, (\mu_0^{(n)}, \mu_1^{(n)})\}$. The core idea of GeONet is to learn the KKT optimality condition~\eqref{eqn:benamou-brenier_kkt} for the Benamou-Brenier problem. Since~\eqref{eqn:benamou-brenier_kkt} is derived to ensure the zero duality gap between the primal and dual dynamic OT problems, solving the Wasserstein geodesic requires us to introduce two sets of neural networks that train the coupled PDEs simultaneously. Specifically, we model the operator learning problem as an enhanced version of the unstacked DeepONet architecture~\citep{LuJinPangZhangKarniadakis2021_DeepONet,https://doi.org/10.48550/arxiv.2202.08942} by jointly training three primal networks in~\eqref{eqn:Continuity_sol} and three dual networks in~\eqref{eqn:HJ_sol} as follows:
%\vspace{-3mm}
\begin{align}
\label{eqn:Continuity_sol}
\mathcal{C}(\mu_0, \mu_1) & (x,t, \phi)  = \sum_{k=1}^p \mathcal{B}_{k}^{0,\text{cty}} (\mu_0, \theta^{0,\text{cty}}) \\ & \cdot  \mathcal{B}_{k}^{1,\text{cty}} (\mu_1,\theta^{1,\text{cty}}) \cdot  \mathcal{T}_{k}^{\text{cty}} (x,t, \xi^{\text{cty}}), \\
\label{eqn:HJ_sol}
\mathcal{H}(\mu_0, \mu_1) & (x,t,\psi)  = \sum_{k=1}^p \mathcal{B}_{k}^{0,\text{HJ}} (\mu_0, \theta^{0,\text{HJ}}) \\ & \cdot \mathcal{B}_{k}^{1,\text{HJ}} (\mu_1, \theta^{1,\text{HJ}}) \cdot \mathcal{T}_{k}^{\text{HJ}} (x,t, \xi^{\text{HJ}}),
\end{align}
where $\mathcal{B}^{j,\text{cty}} (\mu_j(x_1), \dots, \mu_j(x_m), \theta^{j,\text{cty}}) : \mathbb{R}^{m}  \times \Theta \rightarrow \mathbb{R}^p$ and $\mathcal{B}^{j,\text{HJ}} (\mu_j(x_1), \dots, \mu_j(x_m), \theta^{j,\text{HJ}}) : \mathbb{R}^{m}  \times \Theta \rightarrow \mathbb{R}^p$ are \emph{branch} neural networks taking $m$-discretized input of initial and terminal density values at $j = 0$ and $j = 1$ respectively, and $\mathcal{T}^{\text{cty}} (x,t, \xi^{\text{cty}}) : \mathbb{R}^d \times [0,1] \times \Xi \rightarrow \mathbb{R}^p$ and $\mathcal{T}^{\text{HJ}} (x,t, \xi^{\text{HJ}}) : \mathbb{R}^d \times [0,1] \times \Xi \rightarrow \mathbb{R}^p$ are \emph{trunk} neural networks taking spatial and temporal inputs (cf. Appendix~\ref{app:sec:DeepONets} for more details on DeepONet models). Here $\Theta$ and $\Xi$ are finite-dimensional parameter spaces, and $p$ is the output dimension of the branch and truck networks. Denote parameter concatenations $\phi := (\theta^{0,\text{cty}}, \theta^{1,\text{cty}}, \xi^{\text{cty}})$ and $\psi := (\theta^{0,\text{HJ}}, \theta^{1,\text{HJ}}, \xi^{\text{HJ}} )$. Then the primal operator network $\mathcal{C}_{\phi}(x,t,\mu_0, \mu_1) := \mathcal{C}(x,t,\mu_0,\mu_1, \phi)$ for $\phi \in \Theta \times \Theta \times \Xi$ acts as an approximate solution to the continuity equation, hence the true geodesic $\Gamma^{\dag}(x, t, \mu_0(x), \mu_1(x)) := \mu_t(x) = \mu(x,t)$, while the dual operator network $\mathcal{H}_{\psi}(x,t,\mu_0,\mu_1)$ for $\psi \in \Theta \times \Theta \times \Xi$ corresponds to that of the associated Hamilton-Jacobi equation. The architecture of GeONet is shown in Figure~\ref{fig:GeONet_diagram}. 

To train the GeONet defined in~\eqref{eqn:Continuity_sol} and~\eqref{eqn:HJ_sol}, we minimize the empirical loss function corresponding to the system of primal-dual PDEs and boundary residuals in~\eqref{eqn:benamou-brenier_kkt} over the parameter space $\Theta \times \Theta \times \Xi$:
\begin{equation}
\label{eqn:GeONet_loss}
\phi^*, \psi^* \ \ \ = \ \ \ \argmin_{\phi, \psi \in \Theta \times \Theta \times \Xi} \ \ \    \mathcal{L}_{\text{cty}} + \mathcal{L}_{\text{HJ}} + \mathcal{L}_{\text{BC}},  
\end{equation}

%seek a neural network operator $\Upsilon$ that approximates a nontrivial function $ (x, t, \mu_0|_x, \mu_1|_x) \rightarrow \Upsilon^{\dag}(x, t, \mu_0|_x, \mu_1|_x)$ that is the geodesic solution. Let $\Omega \subseteq \mathbb{R}^d$ be a closed domain in Euclidean space, and let $[0,1]$ denote a time interval. We construct two neural networks  

%which we achieve by solving the PDE system of~\eqref{eqn:benamou-brenier_kkt}. This neural operator coheres to the associated physical laws instilled by these equations, consequently ensuring accuracy and regularization, while having no need for geodesic data between initial and target distributions $\mu_0, \mu_1$. Our method is generalized and allows us to feed in a diverse pair $(\mu_0, \mu_1)$ from a distribution family $\Pi$ evaluated at particular $\{x_1, \hdots, x_N\}, x_i \in \mathbb{R}^d$. The geodesic solution is produced, which can be evaluated at any $(x^*,t^*), t^* \in [0,1]$. No retraining or re-computation is needed for new choice of $\mu_0, \mu_1$, unlike with existing methods.

%We begin with data samples from a collection $\{(\mu_0^{[1]}, \mu_1^{[1]}), \hdots, (\mu_0^{[n]}, \mu_1^{[n]})\}$ evaluated over known locations $x \in \Omega$. $x$ can be sampled among its domain, typically being $U(a_1, b_1) \times \hdots \times U(a_d, b_d)$. $x$ is sampled $N$ times per pair, in which pair $(\mu_0, \mu_1)$ is evaluated at these locations. Sampling in the case of images is discussed in section ~\eqref{eqn:CIFAR}.

\begin{strip}
\vspace{-5mm}
\begin{align}
\label{eqn:GeONet_loss_CE}
 \mathcal{L}_{\text{cty},i} & =    \frac{\alpha_{1}}{N} ||  \frac{ \partial}{\partial t} \mathcal{C}_{\phi,i} (x, t)  + \div ( \mathcal{C}_{\phi,i} (x, t) \nabla \mathcal{H}_{\psi,i } (x, t) )  ||_{L^2(\Omega \times (0,1))}^2,  \\
\label{eqn:GeONet_loss_HJ}
\mathcal{L}_{\text{HJ},i} & =  \frac{\alpha_{2}}{N} 
  ||  \frac{\partial}{\partial t} \mathcal{H}_{\psi,i}(x, t)   + \frac{1}{2} || \nabla \mathcal{H}_{\psi,i}(x, t) ||_2^2  ||_{L^2(\Omega \times (0,1))}^2, \\ 
\label{eqn:GeONet_loss_BC}
\mathcal{L}_{\text{BC},i} & = \frac{\beta_{0}}{N} || \mathcal{C}_{\phi,i}(x, 0)  - \mu_{0}^{(i)} ||_{L^2(\Omega)}^2 + \frac{\beta_{1}}{N} || \mathcal{C}_{\phi,i}(x, 1)  - \mu_{1}^{(i)}  ||_{L^2(\Omega)}^2.
\end{align}
\end{strip}
where $\mathcal{L}_{\text{cty}} = \sum_{i=1}^n \mathcal{L}_{\text{cty},i}$, $\mathcal{L}_{\text{HJ}} = \sum_{i=1}^n \mathcal{L}_{\text{HJ},i}$, $\mathcal{L}_{\text{BC}} = \sum_{i=1}^n \mathcal{L}_{\text{BC},i}$, and
Here, $\mathcal{C}_{\phi,i}(x, t) := \mathcal{C}_\phi(x, t, \mu_0^{(i)}(x), \mu_1^{(i)}(x))$ and $\mathcal{C}_{\phi,t,i}(x) := \mathcal{C}_\phi(x, t, \mu_0^{(i)}(x), \mu_1^{(i)}(x))$ denote the evaluation of neural network $\mathcal{C}_{\phi}$ over the $i$-th distribution of the $n$ training data at space location $x$ and time point $t$. The same notation applies for the Hamilton-Jacobi neural networks. $\mathcal{L}_{\text{cty}}$ is the loss component in which the continuity equation is satisfied, and $\mathcal{L}_{\text{HJ}}$ is the Hamilton-Jacobi equation loss component. The boundary conditions are incorporated in the $\mathcal{L}_{\text{BC}}$ term, and $\alpha_1, \alpha_2, \beta_0, \beta_1$ are weights for the strength to impose the physics-informed loss. Automatic differentiation of our GeONet involves differentiating the coupled DeepONet architecture (cf. Figure~\ref{fig:GeONet_diagram}) in order to compute the physics-informed loss terms.

One iterate of our training procedure is as follows. We first select a collection of $N$ indices $\mathcal{I}$ from $1$ to $n$ for which of the $n$ distributions are to be evaluated, with possible repeats. For the physics terms, following~\citep{RAISSI2019686}, we utilize a \emph{collocation} procedure as follows. We sample $N$ pairs $(x,t)$ randomly and uniformly within the bounded domain $\Omega \times [0,1]$. These pairs are resampled during each training iteration in our method. Then, we evaluate the continuity and Hamilton-Jacobi residuals displayed in ~\eqref{eqn:GeONet_loss_CE} and~\eqref{eqn:GeONet_loss_HJ} at such sampled values, in which the loss is subsequently minimized with the corresponding indices in $\mathcal{I}$, making the norms approximated as discrete. For branch input, we take equispaced locations $x_1, \hdots, x_m$ within  $\Omega$, a bounded domain in $\mathbb{R}^d$, typically a hypercube $\tilde{\Omega}$. Then the $N$ branch locations are evaluated among $\tilde{\Omega}$ as well for the BC loss.

%we use a single dual neural network $\Lambda_\xi$ for all training data pairs $(\mu_0^{(i)}, \mu_1^{(i)})_{i=1}^n$. This solution will generally not solve the Hamilton-Jacobi equation belonging to any $(\mu_0^{(i)}, \mu_1^{(i)})$. 

%Data can be concatenated and need not be broken up in the training process. For example, denote $\Lambda_{\xi,i}$ a Hamilton-Jacobi approximate solution for a single pair $(\mu_0^{[i]}, \mu_1^{[i]})$. This solution will generally not solve a Hamilton-Jacobi equation belonging to $(\mu_0^{[j]}, \mu_1^{[j]})$ for $j \neq i$; however, this discrepancy does not need to be resolved in the training procedure and all $n$ Hamilton-Jacobi solutions can be trained with the same neural network $\Lambda_{\xi}$. The addition of the initial and final distributions as neural network inputs allows this. No retraining is needed for new inputs.

\begin{comment}
It is worth highlighting that GeONet is mesh invariant in the sense that geodesics can be evaluated along any location of space and time and no output mesh is needed. The uniform collocation procedure ensures all areas in space and time of the geodesic are considered in training. In contrast, DeepONets do indeed take sample points along some predetermined grid; however, the geodesic itself is perfectly mesh invariant and can be evaluated at any point in space and time. Only the neural network input for the branches scales in length as the initial condition input length scales.
\end{comment}

{\bf Modified multi-layer perceptron (MLP).} A modified MLP architecture as described in~\citep{DBLP:journals/corr/abs-2103-10974} has been shown to have great ability in improving performance for PINNs and physics-informed DeepONets. We elaborate on this architecture in Appendix~\ref{Modified_mlp} and describe our empirical findings with this modified MLP for GeONet in Appendix~\ref{Training and performance}.

{\bf Fourier feature architecture.} An additional augmented architecture is that of the Fourier feature, useful for input data exhibiting fine features, i.e., a lack of spatial differentiability. Spatial-temporal data is transformed using a Fourier mapping. We illustrate in this architecture in~\ref{Fourier_feature}.

%We empirically found success with this modified MLP on GeONet, discussing observed benefits in Appendix~\ref{Training and performance}. We elaborate on this architecture in Appendix~\ref{Modified_mlp}.

%We denote superscript $k$ corresponding to distribution in $\{1,\hdots,n\}$. $d$ corresponds to the spatial dimension of $\Omega \subseteq \mathbb{R}^d$, and $\ell$ corresponds to the particular collocation point, taking values in $\{1, \hdots, N\}$. In the general case, data can be sampled uniformly from within $\Omega \times [0,1]$. For images, spatial data $ \{(x_{1,\ell}^k, \hdots, x_{d, \ell}^k) \}$ is sampled from an equispaced mesh $\Omega^e$. The algorithm requires knowledge in which the distribution pair $(\mu_0^{[i]}, \mu_1^{[i]})$ is evaluated, which can be assigned arbitrarily in the case of images. Data is stored in vectors $X_1, \hdots, X_d, T, U_0, U_1$.

%Our training procedure is conveyed through lines (7)-(13) of the algorithm. We compute the PDE residuals of the continuity and Hamilton-Jacobi equations, as well as error among the boundary conditions. The discrete $|| \cdot ||_{L^2(\Omega \times (0,1))}^2$ norm is taken of these residuals, which we define to be $
%| | f ||_{L^2(\Omega \times (0,1))}^2 \overset{\Delta}{=} \sum_{i} ( f( x_{1,i}, \hdots, x_{d, i}, t_{i} ) )^2 $ . Here, $(x_{1,i}, \hdots, x_{d,i}, t_{i} ), i \in \{1, \hdots, (N)(n)\}$ is the entirety of the collocation points sampled from $U(a_1, b_1) \times \hdots \times U(a_d, b_d) \times U(0,1)$. We set $\Omega = [a_1, b_1] \times \hdots \times [a_d, b_d]$. 

{\bf Entropic regularization.} Our GeONet is compatible with entropic regularization, which is related to the Schr\"odinger bridge problem and stochastic control~\citep{ChenGeorgiouPavon2016}. In Appendix~\ref{app:sec:Entropic_regularization}, we propose the \emph{entropic-regularized GeONet} (ER-GeONet), which learns a similar system of KKT conditions for the optimization as in~\eqref{eqn:benamou-brenier_kkt}. In the zero-noise limit as the entropic regularization parameter $\varepsilon \downarrow 0$, the solution of the optimal entropic interpolating flow converges to solution of the Benamou-Brenier problem~\eqref{eqn:benamou-brenier_formula} in the sense of the method of vanishing viscosity~\citep{Mikami2004,evans2010}.

\section{Numeric experiments}
\label{sec:experiments}

In this section, we perform simulation studies and a real-data example to demonstrate that GeONet can handle inputs as both continuous densities and discrete point clouds (normalized as empirical probability distributions).

%In this section, we conduct experiments to test the performance of GeONet on synthetic Gaussian mixtures and CIFAR-10 data. Gaussian mixtures are a benchmark due to their capability of being smooth density approximators, and CIFAR-10 provides an example of our method on real image data.

\subsection{Input as continuous density: Gaussian mixture distributions}
\label{subsec:gaussian_mixtures}

\begin{figure*}[t]
  \centering
  \includegraphics[scale=0.53]{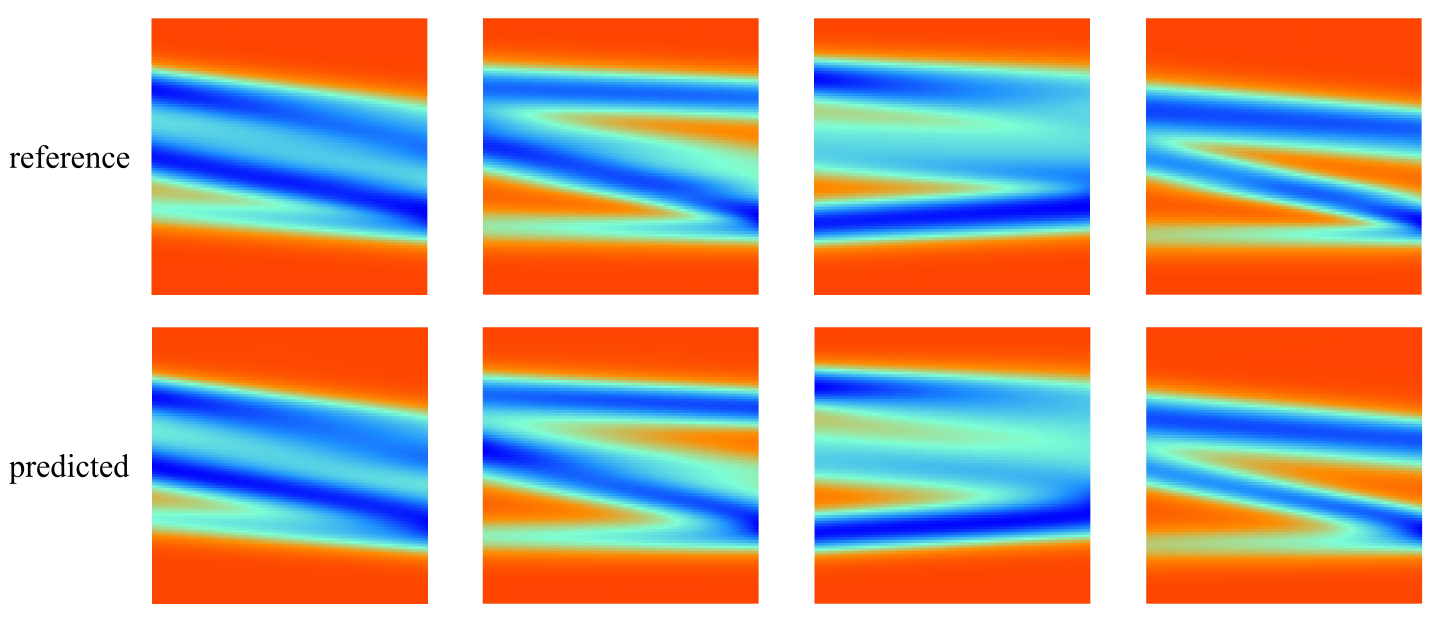}
  \vspace{-2mm}
  \caption{Four geodesics predicted by GeONet with reference geodesics computed by POT on test univariate Gaussian mixture distribution pairs with $k_0 = k_1 = 6$. The reference serves as a close approximation to the true geodesic. The vertical axis is space and the horizontal axis is time.}
  \vspace{-5mm}
  \label{fig:GeONet_gaussian_mixture}
\end{figure*}

Since finite mixture distributions are powerful universal approximators for continuous probability density functions~\citep{doi:10.1080/25742558.2020.1750861}, we first deploy GeONet on Gaussian mixture distributions over domains of varying dimensions. We learn the Wasserstein geodesic mapping between two distributions of the form $\mu_j(x) = \sum_{i=1}^{k_j} \pi_i \mathcal{N}(x | u_i , \Sigma_i)$  \text{subject to} $ \sum_{i=1}^{k_j} \pi_i = 1$,
where $j \in \{0,1\}$ corresponds to initial and terminal distributions $\mu_0, \mu_1$, and $k_j$ denotes the number of components in the mixture. Here $u_i$ and $\Sigma_i$ are the mean vectors and covariance matrices of individual Gaussian components respectively. Due to the space limit, we defer simulation setups, model training details and error metrics to Appendices~\ref{Hyperparameter_settings},~\ref{Training and performance} and~\ref{app:error_metrics}, respectively.

\begin{comment}
\begin{figure}[h]
  \centering
  %\includegraphics[scale=0.55]{Gaussians 8.pdf}
  \includegraphics[scale=0.58]{GeONet_univariate_samples_2.pdf}
  \vspace{-5mm}
  \caption{Four unique geodesics predicted by GeONet with reference geodesics computed by POT on test univariate Gaussian mixture distribution pairs with $k_0 = k_1 = 6$. The reference serves as a close approximation to the true geodesic due to entropic regularization. The vertical axis is space and the horizontal axis is time. \vspace{-8mm}}
  \label{fig:GeONet_gaussian_mixture}
\end{figure}
\end{comment}

\begin{comment}
\begin{figure}[h]
  \centering
  %\includegraphics[scale=0.55]{Gaussians 8.pdf}
  \includegraphics[scale=0.56]{GeONet_bivariate_samples_8.pdf}
  \caption{Geodesics predicted by GeONet on bivariate Gaussians over a square domain. The top of each pair is the reference solution computed by POT, and the bottom is GeONet.  \vspace{-3mm}}
  \label{fig:GeONet_gaussian_mixture_bivariate}
\end{figure}
\end{comment}

We examine errors in regard to an identity geodesic (i.e., $\mu_0 = \mu_1$), a random test pairing, and an out-of-distribution (OOD) pairing. The mesh-invariant nature of the output of GeONet allows zero-shot super-resolution for adapting low-resolution data into high-resolution geodesics, which includes initial data at $t=0,1$, while traditional OT solvers and non-operator learning based methods have no ability to do this, as they are confined to the original mesh. Thus, we also include a random test pairing on higher resolution than training data. The result is reported in Table~\ref{tab:GeONet_gaussian_mixture}.

%Since $\int_{\Omega} | \mu| \rd x = 1$ for all time points, the $L^1$ error is also a relative error, a meaningful metric essentially corresponding to the percentage error between the neural operator geodesic and the reference. %Relative $L^2$ error is a common metric among comparisons, but we do not have previous geodesic results to compare, in which our proposed metric is more suitable.
%We examine errors in regard to an identity geodesic (i.e., $\mu_0 = \mu_1$), a random test pairing, a random test high-resolution pairing, and an out-of-distribution (OOD) pairing. 

\begin{table*}[t]
  \vspace{-3mm}
  \caption{$L^1$ error of GeONet on 50 test data of univariate and bivariate Gaussian mixtures. We compute errors on cases of the identity geodesic, a random pairing in which $\mu_0 \neq \mu_1$, high-resolution random pairings refined to $200$ and $75 \times 75$ resolutions in the 1D and 2D cases respectively, and out-of-distribution examples. We report the means and standard deviations as a percentage, making all values multiplied by $10^{-2}$ by those of the table.\vspace{0mm}}
  % In the second part, we train and test upon $k_0 = k_1 = 5$, $\pi_i = 0.2$ for all $i$, with the same loss coefficients. 
  \label{tab:GeONet_gaussian_mixture}
  \centering
  \begin{tabular}[b]{
    l
    S[table-format = 3]
    S[table-format = 2]
    S[table-format = 1.3]
    S[table-format = -2.2]
    S[table-format = 1.3]
    S[table-format = 1.3]
    S[table-format = 2.2]
    }
    \toprule
    \multicolumn{1}{c}{} & 
    \multicolumn{5}{c}{GeONet $L^1$ error for Gaussian mixtures}\\
    \cmidrule(lr){2-6}        
    \textbf{Experiment \ \ } & {$\bm{ t=0 }$} & {$\bm{  t=0.25  }$} & {$\bm{ t=0.5  }$} & {$\bm{ t=0.75  }$} & {$\bm{ t =1 }$}  \\
    \midrule
    \text{1D identity} & 
    {$2.67 \pm 0.750$} & 
    {$2.85 \pm 0.912$} &
    {$3.04 \pm 1.02$} & 
    {$2.86 \pm 0.898$} &
    {$2.63 \pm 0.696$} \\
    \text{1D random}  &
    {$4.92 \pm 2.00$}  & 
    {$5.43 \pm 3.02$}  & 
    {$5.76 \pm 3.56$} & 
    {$5.26 \pm 3.25$} & 
    {$4.65 \pm 1.50$} \\
    \text{1D high-res.}  &
    {$4.76 \pm 1.53$}  & 
    {$5.49 \pm 3.00$}  & 
    {$6.01 \pm 3.53$} & 
    {$5.59 \pm 2.99$} & 
    {$4.77 \pm 1.49$} \\
    \text{1D OOD} & 
    {$12.9 \pm 4.13$} &
    {$14.3 \pm 5.35$} &  
    {$16.4 \pm 6.01$} &
    {$14.9 \pm 5.48$} &
    {$12.3 \pm 3.94$} \\
    \midrule
    \text{2D identity} & 
    {$6.50 \pm 1.15$} & 
    {$7.68 \pm 0.915$} &
    {$7.69 \pm 0.924$} & 
    {$7.70 \pm 0.889$} &
    {$6.42 \pm 1.11$} \\
    \text{2D random}  &
    {$6.59 \pm 1.01$} & 
    {$7.10 \pm 0.869$} &
    {$7.13 \pm 0.892$} & 
    {$7.04 \pm 0.780$} &
    {$6.33 \pm 0.835$} \\
    \text{2D high-res.}  &
    {$6.66 \pm 0.766$} & 
    {$7.71 \pm 1.26$} &
    {$7.88 \pm 1.21$} & 
    {$7.59 \pm 0.979$} &
    {$6.29 \pm 0.723$} \\
    \text{2D OOD} & 
    {$7.15 \pm 0.985$} &
    {$7.82 \pm 1.04$} &  
    {$8.14 \pm 1.33$} &
    {$7.96 \pm 1.30$} &
    {$7.14 \pm 0.882$} \\
    
    %\hline
    %\text{Identity $k_0 = k_1 = 5$} & 
    %{$0.23 \pm 0.23$} & 
    %{$1.7 \pm 0.67$} &
    %{$2.6 \pm 1.1$} & 
    %{$1.8 \pm 0.74$} &
    %{$0.22 \pm 0.21$} \\
    %\text{Generic $k_0 = k_1 = 5$} & 
    %{$0.22 \pm 0.15$} & 
    %{$1.7 \pm 0.92$} &
    %{$2.7 \pm 1.5$} & 
    %{$1.7 \pm 0.76$} &
    %{$0.20 \pm 0.13$} \\
    \bottomrule
\end{tabular}
\vspace{0mm}
\end{table*}

\subsection{Input as point clouds: Gaussian mixture distributions}
\label{empirical_Gaussians}

\begin{figure}[h]
  \centering
  \includegraphics[scale=0.55]{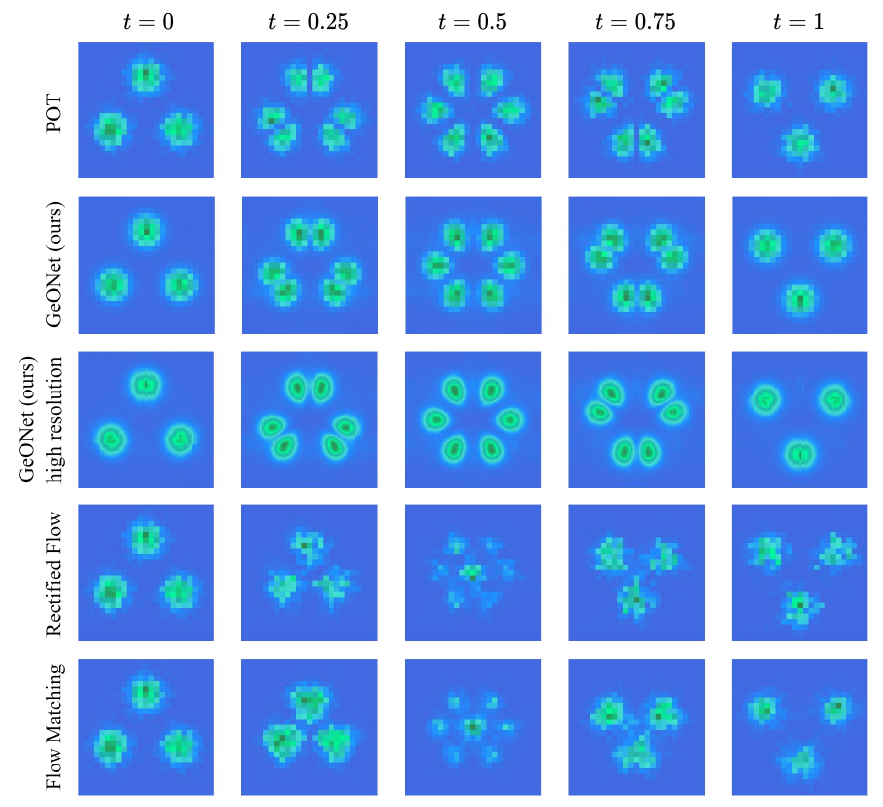}
  \vspace{-4mm}
  \caption{We compare to GeONet to the alternative methodology in a discrete setting, using POT as ground truth. GeONet is the only method among the comparison which encapsulates the geodesic behaviour among the translocation of points. \vspace{-2mm}}
  \label{fig:GeONet_pointclouds}
\end{figure}

GeONet is also suited for continuous densities made discrete. In scenarios with access to point clouds of data, we may use GeONet with discrete data made into empirical distributions. We test GeONet an an example Gaussian discrete data. Discrete particles in $\Omega \subseteq \mathbb{R}^2$ are sampled from Gaussian data, as encompassed in ~\citep{liu2022flow}. The sampled particles are represented by empirical densities, in which we compare upon the transition of densities in the non-particle setting using POT as a baseline. The process of turning sampled particles into an empirical density can be reversed by sampling particles according to the densities, which in its most simplified form is placing the particles exactly along the mesh, with the number corresponding to rounded density value. The result is reported in Table~\ref{tab:GeONet_gaussian_mixture_discrete} and an estimated geodesic example is shown in Figure~\ref{fig:GeONet_pointclouds}. We observe that conditional flow matching (CFM)~\citep{tong2023improving} and rectified flow (RF)~\citep{liu2022flow} have 3-4 times comparably larger estimation errors than GeONet, except for the initial time $t = 0$, because this initial data is given and learned directly for RF and CFM. GeONet is the only framework among the comparison which encapsulates the geodesic behavior to a considerable degree. Second, RF and CFM have the same fixed resolution as the input probability distribution pairing, while GeONet can be smoothed out for estimating the density flows on higher resolution than the input pairing (cf. the third row in Figure~\ref{fig:GeONet_pointclouds}).

%\setcitestyle{numbers}

\begin{table*}[h]
  \caption{$L^1$ error between GeONet, the conditional flow matching (CFM) library's optimal transport solver~\cite{tong2023improving}, and rectified flow (RF)~\cite{liu2022flow}, using POT again as a baseline for comparison. All values are multiplied by $10^{-2}$ to those of the table.\vspace{0mm}}
  % In the second part, we train and test upon $k_0 = k_1 = 5$, $\pi_i = 0.2$ for all $i$, with the same loss coefficients. 
  \label{tab:GeONet_gaussian_mixture_discrete}
  \centering
  \begin{tabular}[b]{
    l
    S[table-format = 3]
    S[table-format = 2]
    S[table-format = 1.3]
    S[table-format = -2.2]
    S[table-format = 1.3]
    S[table-format = 1.3]
    S[table-format = 2.2]
    }
    \toprule
    \multicolumn{1}{c}{} & 
    \multicolumn{5}{c}{$L^1$ comparison error on 2D Gaussian mixture point clouds}\\
    \cmidrule(lr){2-6}        
    \textbf{Experiment \ \ } & {$\bm{ t=0 }$} & {$\bm{  t=0.25  }$} & {$\bm{ t=0.5  }$} & {$\bm{ t=0.75  }$} & {$\bm{ t =1 }$}  \\
    \midrule
    \text{GeONet}  &
    {$22.9 \pm 1.08$}  & 
    {$28.8 \pm 1.01$}  & 
    {$30.0 \pm 1.10$} & 
    {$29.6 \pm 0.877$} & 
    {$22.6 \pm 1.02$} \\
    \text{CFM}  &
    {$0.0 \pm 0.0$} & 
    {$94.1 \pm 3.68$} &
    {$98.9 \pm 2.41$} & 
    {$91.8 \pm 4.15$} &
    {$75.9 \pm 3.77$} \\
    \text{RF}  &
    {$0.0 \pm 0.0$} & 
    {$103 \pm 2.48$} &
    {$112 \pm 3.61$} & 
    {$112 \pm 5.03$} &
    {$91.3 \pm 3.79$} \\
    
    %\hline
    %\text{Identity $k_0 = k_1 = 5$} & 
    %{$0.23 \pm 0.23$} & 
    %{$1.7 \pm 0.67$} &
    %{$2.6 \pm 1.1$} & 
    %{$1.8 \pm 0.74$} &
    %{$0.22 \pm 0.21$} \\
    %\text{Generic $k_0 = k_1 = 5$} & 
    %{$0.22 \pm 0.15$} & 
    %{$1.7 \pm 0.92$} &
    %{$2.7 \pm 1.5$} & 
    %{$1.7 \pm 0.76$} &
    %{$0.20 \pm 0.13$} \\
    \bottomrule
\end{tabular}
%\vspace{-6mm}
\end{table*}

\setcitestyle{names}

\subsection{A real data application}
\label{eqn:MNIST}

\begin{figure*}[!t]
\vspace{0mm}
  \centering
  \includegraphics[scale=0.78]{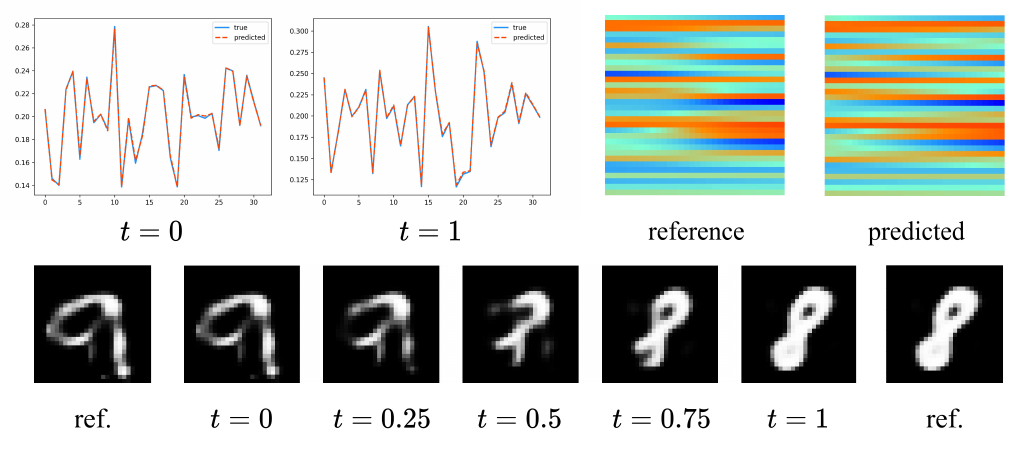}
  \caption{Beginning from top left and going clockwise, we display the initial conditions in the encoded space, the geodesics in the encoded space, and the decoded geodesics as $28 \times 28$ images. (a) and (b) correspond to two unique pairings.}
  \label{fig:GeONet_MNIST}
  \vspace{-1mm}
\end{figure*}

Our next experiment was upon the MNIST dataset of $28 \times 28$ images of single-digit numbers. It is difficult for GeONet to capture the geodesics between digits: MNIST resembles jump-discontinuous data, and relatively piecewise constant otherwise, which is troublesome for the physics-informed term. To remedy our problems with MNIST, we use a pretrained autoencoder to encode the MNIST digits into a low-dimensional representation $v \in \mathbb{R}^{32}$ with an encoder $\Phi$ and a decoder $\Phi^{-1} : v \rightarrow \mathbb{R}^{28} \times \mathbb{R}^{28}$ mapping the encoded representation into newly-formed digits resembling that which was fed into the encoder. We institute GeONet upon the encoded representations, learning the geodesic between highly irregular encoded data. Table~\ref{tab:GeONet_CIFAR-10} reports the $L^1$ errors for geodesic estimated in the encoded space and recovered images in the ambient space. As expected, the ambient-space error is much larger than the encoded-space error, meaning that the geodesics in the encoded space and ambient image space do not coincide. Figure~\ref{fig:GeONet_MNIST} shows the learnt geodesics in the encoded space and decoded images on the geodesics. 

\subsection{Runtime comparison}

\begin{comment}
%------------------------------------------
\begin{wrapfigure}{r}{0.49\textwidth}
	
\vspace{-10.4pt}
  \centering
    \includegraphics[width=0.42\textwidth]{geonet_cost_graph_1.pdf}
    \vspace{-0.5mm}
  \caption{Comparison of log-runtime in seconds on Gaussian mixtures. The y-axis is the log of time, the x-axis is the log of grid length in one dimension. We total over 50 density pairs.}
  \label{fig:POT_GeONet_Comparison}
\end{wrapfigure}
%------------------------------------------
\end{comment}

\begin{figure*}
  \centering
  \includegraphics[scale=0.58]{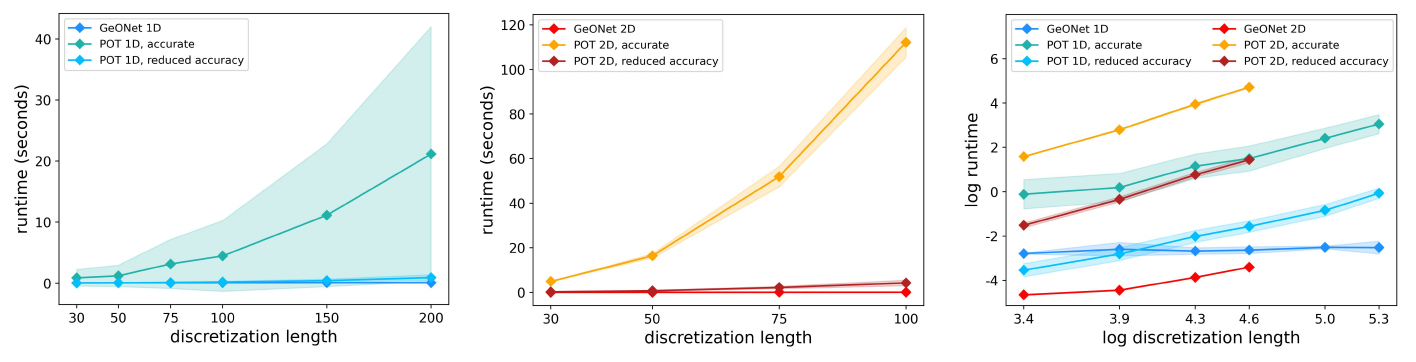}
  \vspace{0mm}
  \caption{We compare to GeONet to the classical POT library on 1D and 2D Gaussians in terms of mean and standard deviations of runtime on both an unmodified scale as well as one that is log-log using discretization length in one dimension as the x-axis, taken over 30 pairs. We use 20 time steps for 1D and 5 for 2D. Finer meshes are omitted for 2D for computational reasonableness. \vspace{-3mm}}
  \label{fig:GeONet_runtime}
\end{figure*}

\begin{comment}
Inference of GeONet is instantaneous, a feature advantageous for many pairs and high resolution images, especially if geodesics are needed quickly. POT is greatly encumbered by a fine mesh. Figure~\ref{fig:POT_GeONet_Comparison} (right) demonstrates an inference-stage runtime comparison of GeONet to the POT algorithms on Gaussian mixtures considered in Section~\ref{subsec:gaussian_mixtures}. The POT algorithm was minimally modified, only so that the pertinent distributions could be taken as input. The linear pattern of Figure~\ref{fig:POT_GeONet_Comparison} (right) on log-scale suggests the computational improvement of GeONet over POT is of the orders of magnitude. In Figure~\ref{fig:POT_GeONet_Comparison}, the number of time discretization is 20 for univariate, and 5 for bivariate. The time to create the data is not considered in the figure. More training details can be found in Appendix~\ref{Hyperparameter_settings}.
\end{comment}
Our method is highlighted by the fact that it is near instantaneous: it is highly suitable when many geodesics are needed quickly, or over fine meshes. Traditional optimal transport solvers are greatly encumbered when evaluated over a fine grid, but the output mesh-invariant nature of GeONet bypasses this. In Figure~\ref{fig:GeONet_runtime}, we illustrate GeONet versus POT, a traditional OT library. GeONet greatly outperforms POT for fine grids, especially if POT is used to compute an accurate solution. Even when POT is used to equivalent accuracy, GeONet still outperforms, most illustrated in the log-log plot. The log-log plot also demonstrates that our method speeds computation up to orders of magnitude. We restrict the accuracy of POT by employing a stopping threshold of $0.5$ for 1D and $10.0$ for 2D. We found these choices were comparable to GeONet, remarking a threshold of $10.0$ in the 2D case is sufficiently large so that even larger thresholds have limited effect on error.

\subsection{Out-of-distribution generalization}

To examine the generalization performance of GeONet under distribution shifts, we consider a setup which is out-of-distribution (OOD) at test time. To do this, we alter the variances to be lower than in the training setup. Changing the number of Gaussians in the mixture at test time does not yield a significant enough of an alteration to the experiment. For univariate Gaussians, variance was among $[0.3,0.4]$ instead of the $[0.5,0.6]$ in training. For bivariate Gaussians, we take variances in $[0.25,0.3]$ and $[0.65,0.8]$, six of each in each mixture. Empirically, we found OOD generalization had lower error in this experiment than in the univariate case, which may be because the support of areas of low variance have relatively low measure. This bivariate error does indeed scale greater as variance decreases even more. We remark the lower variance generally yields a more sophisticated task in learning the geodesic, and error even trained upon such data is higher than with high variance.

\begin{figure}[h]
\centering
\includegraphics[scale=0.6]{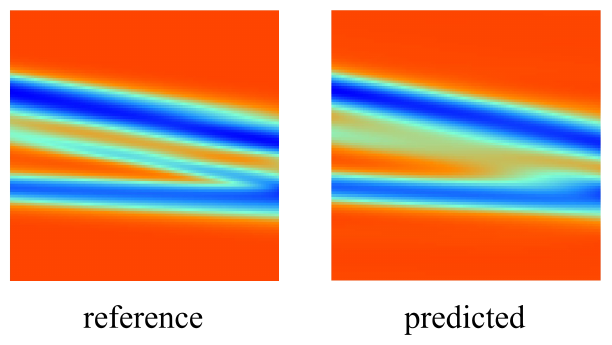}
\caption{An illustration of an OOD univariate Gaussian mixture geodesic at test time.}
\label{fig:OOD_Univariate}
\end{figure}

\subsection{Limitations}
%GeONet branch input scales with respect to the spatial dimension, quickly requiring large amounts of input input even in moderately high dimensions. A remedy for this problem is to encode the data in a low-dimensional representation similar to what we have seen in the MNIST experiment. On the other hand, most traditional geodesic solvers focus on functions over one or two dimensions, while GeONet serves as a secondary option which can indeed extend to any dimension, setting computational rigor and accuracy aside. Furthermore, GeONet requires predetermined locations in which branch input is evaluated; however, this setup is necessary due to the pair of initial conditions.

There are several limitations we would like to address. First, GeONet's branch network input exponentially increases in spatial dimension, necessitating extensive input data even in moderately high-dimensional scenarios. One strategy to mitigate this is through leveraging low-dimensional data representations as in the MNIST experiment. While traditional geodesic solvers primarily handle one or two dimensions, GeONet offers a versatile alternative, accommodating any dimension at the cost of potential computational precision. Second, GeONet mandates predetermined evaluation points for branch input, a requisite grounded in the pairing of initial conditions. It is of interest to extend GeONet to include training input data pair on different resolutions. Third, given the regularity of the OT problem~\citep{hutter2021minimax,Caffarelli_1996}, developing a generalization error bound for assessing the predictive risk of GeONet is an important future work. Finally, the dynamical OT problem is closely connected to the mean-field planning with an extra interaction term~\citep{FU2023112346}. It would be interesting to extend the current operator learning perspective to such problems.

%\subsection{Limitations}
%We discuss limitations in Appendix~\ref{Training and performance}.

%\subsection{Limitations}

%A key limitation is training time. Also, there is a requirement of input function values uniformly evaluated along the same points in space for each function. Additionally, we found the empirical error relatively high to the baseline error for the identity geodesic. Physics-informed DeepONets have found great success over domains $(x,t) \in \mathbb{R} \times \mathbb{R}$. Solving a coupled PDE system over $(x,t) \in \mathbb{R}^2 \times \mathbb{R}$ is what we introduce, and  error for such a case has been less studied.

%\section{Acknowledgments}

%This section is currently omitted for anonymity.

%Comparison of log-scale computer runtime (in seconds) for POT and GeONet over 50 testing pairs with vary mesh sizes.

%The lack of the requirement of geodesic data for training accelerates training, as such data-generation is additionally costly. More collocations are generally needed in high resolution images, similar to the CIFAR-10 case.

\clearpage
\bibliographystyle{plainnat}
\bibliography{neural_operator}

\onecolumn

\title{GeONet: a neural operator for learning the Wasserstein geodesic (Supplementary Material)}
\maketitle 

\appendix

\section{Training algorithm}

\begin{algorithm}
\caption{End-to-end training of GeONet}\label{alg:cap}
\textbf{Input:} data pairs $(\mu_0^{(1)} \mu_1^{(1)}), \hdots, (\mu_0^{(n)}, \mu_1^{(n)})$; batch size $N$;  initialization of the neural network parameters $\phi, \psi \in \Theta \times \Theta \times \Xi$; weight parameters $\alpha_1, \alpha_2, \beta_0, \beta_1$; domain $\Omega$ and branch domain (mesh) $\tilde{\Omega}$.; denote $i \in \{1,\hdots,N\}$.
%\textbf{Initialize:} vectors $X$ and $T$.
\begin{algorithmic}[1]
\While{$\mathcal{L}_{\text{total}}$ has not converged}
\State Independently draw $N$ sample points from $U(\Omega) \times U(0,1)$, $N$ points from $U(\tilde{\Omega})$, and $N$ density pairs from $\{(\mu_0^{(\ell)},\mu_1^{(\ell)})\}_{\ell=1}^n$, possibly repeating.
\State Compute $\Phi_i =  \partial_{t} \mathcal{C}_{\phi,i} + \div(\mathcal{C}_{\phi,i} \nabla \mathcal{H}_{\psi,i} )   $.
\Comment{\texttt{continuity residual}}
\State Compute $\Psi_i =    \partial_{t} \mathcal{H}_{\psi,i} + {1\over2}\| \nabla \mathcal{H}_{\psi,i} \|_2^2  $. \Comment{\texttt{Hamilton-Jacobi residual}}
\State Compute $B_{0,i} =    \mathcal{C}_{\phi,0,i}  - \mu_0^{(i)}(x_{\tilde{\Omega}}^i), \ \ B_{1,i} = \mathcal{C}_{\phi,1,i} - \mu_1^{(i)}(x_{\tilde{\Omega}}^i)    $. \Comment{\texttt{boundary residual}}
\State Compute 
\[
\begin{gathered}
\mathcal{L}_{\text{cty}} =  \frac{\alpha_1}{N} \sum_{i=1}^N \Phi_i^2, \ \ \ \  \mathcal{L}_{\text{HJ}} = \frac{\alpha_2}{N} \sum_{i=1}^N  \Psi_i^2 , \\
\mathcal{L}_{\text{BC}} = \frac{1}{N}  \sum_{i=1}^N (  \beta_0  B_{0,i}^2 + \beta_1  B_{1,i}^2 ),
\end{gathered}
\]
\State Compute $\mathcal{L}_{\text{total}}(\phi, \psi) = \mathcal{L}_{\text{cty}} + \mathcal{L}_{\text{HJ}} + \mathcal{L}_{\text{BC}}$.
\State Minimize $\mathcal{L}_{\text{total}}(\phi, \psi)$ to update $\phi$ and $\psi$. \Comment{\texttt{minimize the loss function}}
\EndWhile
\end{algorithmic}
\end{algorithm}

\section{Derivation of primal-dual optimality conditions for dynamical OT problem}
\label{app:sec:primal-dual_optimality}

The primal-dual analysis is a standard technique in the optimization literature such as in analyzing certain semidefinite programs~\citep{9366690}. Recall the Benamou-Brenier fluid dynamics formulation of the static optimal transport problem
\begin{align}
\label{app:eqn:min_kinetic_energy}
& \min_{(\mu, \vv)}  \int_0^1 \int_{\mathbb{R}^d} {1\over2} || \vv(x, t) ||_2^2 \ \mu(x, t) \ \rd x \ \rd t \\
\label{app:eqn:continuity_equation} 
 & \mbox{subject to}  \ \  \partial_t \mu + \div(\mu \vv) = 0,\\
 \label{app:eqn:boundary_condition}
 &   \mu(\cdot, 0) = \mu_0, \ \  \mu(\cdot, 1) = \mu_1.
\end{align}
Here, equation~\eqref{app:eqn:continuity_equation} is referred to as the \emph{continuity equation} (CE), preserving the unit mass of the density flow $\mu_t = \mu(\cdot, t)$. We write the Lagrangian function for any flow $(\mu_t)_{t \in [0, 1]}$ initializing from $\mu_0$ and terminating at $\mu_1$ as
\begin{equation}
    \label{app:eqn:lagrangian_benamou-brenier}
    \begin{gathered}
    L(\mu, \vv, u) = \int_0^1 \int_{\bR^d} \left[ {1\over2} \|\vv\|_2^2 \mu + \left( \partial_t \mu + \div(\mu \vv) \right) u \right] \; \rd x \; \rd t,
    \end{gathered}
\end{equation}
where $u := u(x, t)$ is the dual variable for (CE). To find the optimal solution $\mu^*$ for the minimum kinetic energy~\eqref{app:eqn:min_kinetic_energy}, we study the saddle point optimization problem
\begin{equation}
    \label{app:eqn:lagrangian_benamou-brenier_saddle_point}
    \min_{(\mu, \vv) \in \text{(CE)}} \max_u L(\mu, \vv, u),
\end{equation}
where the minimization over $(\mu, \vv)$ runs over all flows satisfying (CE) such that $\mu(\cdot, 0) = \mu_0$ and $\mu(\cdot, 1) = \mu_1$. Note that if $\mu \notin \text{(CE)}$, then by scaling with arbitrarily large constant, we see that
\begin{equation}
    \max_u \int_0^1 \int_{\bR^d} \left( \partial_t \mu + \div(\mu \vv) \right) u \; \rd x \; \rd t = + \infty.
\end{equation}
Thus,
\begin{align}
    \min_{(\mu, \vv) \in \text{(CE)}} \int_0^1 \int_{\mathbb{R}^d} {1\over2} || \vv ||_2^2 \mu \ dx \ dt = & \min_{(\mu, \vv)} \max_u L(\mu, \vv, u) \\ \geq & \max_u \min_{(\mu, \vv)} L(\mu, \vv, u),
\end{align}
where the minimization over $(\mu, \vv)$ is unconstrained.  Using integration-by-parts and suitable decay for vanishing boundary as $\|x\|_2 \to \infty$, we have
\begin{align*}
    L(\mu, \vv, u) = & \int_0^1 \int_{\bR^d} \left[ {1\over2} \|\vv\|_2^2 \mu - \mu \partial_t u - \langle \vv, \nabla u \rangle \mu \right] \; \rd x \; \rd t \\
    & \qquad + \int_{\bR^d} \left[ \mu(\cdot,1) u(\cdot,1) - \mu(\cdot,0) u(\cdot,0) \right] \; \rd x.
\end{align*}
Now, we fix $\mu$ and $u$, and minimize $L(\mu, \vv, u)$ over $\vv$. The optimal velocity vector is $\vv^* = \nabla u$, and we have
\begin{equation}
    \max_u \min_{\mu} L(\mu, \vv^*, u) = \int_0^1 \int_{\bR^d} \left[ -\left( {1\over2} \|\nabla u\|_2^2 + \partial_t u \right) \mu \right] \; \rd x \; \rd t  + \int_{\bR^d} \left[ u(\cdot,1) \mu_1 -  u(\cdot,0) \mu_0 \right] \; \rd x,
\end{equation}
for any flow $\mu_t$ satisfying the boundary conditions $\mu(\cdot, 0) = \mu_0$ and $\mu(\cdot, 1) = \mu_1$. If ${1\over2} \|\nabla u\|_2^2 + \partial_t u \neq 0$, then by the same scaling argument above, we have
\begin{equation}
    \min_{\mu} \int_0^1 \int_{\bR^d} \left[ -\left( {1\over2} \|\nabla u\|_2^2 + \partial_t u \right) \mu \right] \; \rd x \; \rd t = -\infty
\end{equation}
because $\mu$ is unconstrained (except for the boundary conditions). Then we deduce that
\begin{equation}
\label{app:eqn:duality}
    \min_{(\mu, \vv) \in \text{(CE)}} \int_0^1 \int_{\mathbb{R}^d} {1\over2} || \vv ||_2^2 \mu \geq \max_{u \in \text{(HJ)}} \left\{ \int_{\bR^d} u(\cdot,1) \mu_1 - \int_{\bR^d} u(\cdot,0) \mu_0 \right\},
\end{equation}
where $u \in \text{(HJ)}$ means that $u$ solves the \emph{Hamilton-Jacobi equation} (HJ)
\begin{equation}
    \label{app:eqn:hamilton-jacobi}
    \partial_t u + {1\over2} \|\nabla u\|_2^2 = 0.
\end{equation}
From~\eqref{app:eqn:duality}, we see that the duality gap is non-negative, and it is equal to zero if and only if $(\mu^*, u^*)$ solves the following system of PDEs
\begin{equation}
    \label{app:eqn:benamou-brenier_kkt}
    \left\{
    \begin{gathered}
      \partial_t \mu + \div(\mu \nabla u) = 0, \ \ \partial_t u + {1\over2} \|\nabla u\|_2^2 = 0, \\
     \mu(\cdot,0) = \mu_0, \ \ \mu(\cdot,1) = \mu_1.
     \end{gathered} \right.
\end{equation}
PDEs in~\eqref{app:eqn:benamou-brenier_kkt} are referred to as the Karush–Kuhn–Tucker (KKT) conditions for the Wasserstein geodesic problem.

\section{Metric geometry structure of the Wasserstein space and geodesic}
\label{app:sec:wasserstein_facts}

In this section, we review some basic facts on metric geometry properties of the Wasserstein space and geodesic. We first discuss the general metric space $(X, d)$, and then specialize to the Wasserstein (metric) space $({\cal P}_p(\bR^d), W_p)$ for $p \geq 1$. Furthermore, we connect to the fluid dynamic formulation of optimal transport. Most of the materials are based on the reference books~\citep{BurageBuragoIvanov2001_MetricGeometry,AmbrosioGigliSavare2008_GradientFlows,sabtanbrogio2015_OT}.

\subsection{General metric space}

\begin{defn}[Absolutely continuous curve]
\label{def:ac_curve}
Let $(X, d)$ be a metric space. A curve $\omega : [0, 1] \to X$ is \emph{absolutely continuous} if there is a function $g \in L^1([0,1])$ such that for all $t_0 < t_1$, we have
\begin{equation}
\label{eqn:ac_curve}
d(\omega(t_0), \omega(t_1)) \leq \int_{t_0}^{t_1} g(\tau) \, \rd \tau.
\end{equation}
Such curves are denoted by $\text{AC}(X)$.
\end{defn}

\begin{defn}[Metric derivative]
\label{def:metric_derivative}
If $\omega : [0,1] \to X$ is a curve in a metric space $(X, d)$, the \emph{metric derivative} of $\omega$ at time $t$ is defined as
\begin{equation}
    \label{eqn:metric_derivative}
    |\omega'|(t) := \lim_{h \to 0} {d(\omega(t+h), \omega(t)) \over |h|},
\end{equation}
if the limit exists.
\end{defn}

The following theorem generalizes the classical Rademacher theorem from a Euclidean space into any metric space in terms of the metric derivative.

\begin{thm}[Rademacher]
\label{thm:rademacher}
If $\omega : [0,1] \to X$ is Lipschitz continuous, then the metric derivative $|\omega'|(t)$ exists for almost every $t \in [0,1]$. In addition, for any $0 \leq t < s \leq 1$, we have
\begin{equation}
    d(\omega(t), \omega(s)) \leq \int_{t}^{s} |\omega'|(\tau) \, \rd \tau.
\end{equation}
\end{thm}

Theorem~\ref{thm:rademacher} tells us that absolutely continuous curve $\omega$ has a metric derivative well-defined almost everywhere, and the ``length" of the curve $\omega$ is bounded by the integral of the metric derivative. Thus, a natural definition of the length of a curve in a general metric space is to take the best approximation over all possible meshes.

\begin{defn}[Curve length]
\label{def:curve_length}
For a curve $\omega : [0,1] \to X$, we define its \emph{length} as
\begin{equation}
    \label{eqn:curve_length}
    \text{Length}(\omega) := \sup \left\{ \sum_{k=0}^{n-1} d(\omega(t_k), \omega(t_{k+1})) : n \geq 1, 0 = t_0 < t_1 < \hdots < t_n = 1 \right\}.
\end{equation}
\end{defn}

Note that if $\omega \in \text{AC}(X)$, then
\begin{equation}
    d(\omega(t_k), \omega(t_{k+1})) \leq \int_{t_k}^{t_{k+1}} g(\tau) \, \rd \tau 
\end{equation}
so that
\begin{equation}
    \text{Length}(\omega) \leq \int_0^1 g(\tau) \, \rd \tau < \infty,
\end{equation}
i.e., the curve $\omega$ is of bounded variation.

\begin{lem}
If $\omega \in \text{AC}(X)$, then
\begin{equation}
    \text{Length}(\omega) = \int_0^1 |\omega'|(\tau) \, \rd \tau.
\end{equation}
\end{lem}

\begin{defn}[Length space and geodesic space]
Let $\omega : [0,1] \to X$ be a curve in $(X, d)$.
\begin{enumerate}
    \item The space $(X, d)$ is a \emph{length space} if
    \begin{equation}
        d(x, y) = \inf \left\{ \text{Length}(\omega) : \omega(0) = x, \omega(1) = y, \omega \in \text{AC}(X) \right\}.
    \end{equation}
    
    \item The space $(X, d)$ is a \emph{geodesic space} if
    \begin{equation}
        d(x, y) = \min \left\{ \text{Length}(\omega) : \omega(0) = x, \omega(1) = y, \omega \in \text{AC}(X) \right\}.
    \end{equation}
\end{enumerate}
\end{defn}

\begin{defn}[Geodesic]
Let $(X, d)$ be a length space. 
\begin{enumerate}
    \item A curve $\omega : [0,1] \to X$ is said to be a \emph{constant-speed geodesic} between $\omega(0)$ and $\omega(1)$ if
    \begin{equation}
        d(\omega(t), \omega(s)) = |t-s| \cdot d(\omega(0), \omega(1)),
    \end{equation}
    for any $t, s \in [0, 1]$.
    
    \item If $(X, d)$ is further a geodesic space, a curve $\omega : [0,1] \to X$ is said to be a \emph{geodesic} between $x_0 \in X$ and $x_1 \in X$ if it minimizes the length among all possible curves such that $\omega(0) = x_0$ and $\omega(1) = x_1$.
\end{enumerate}
\end{defn}
Note that in a geodesic space $(X, d)$, a constant-speed geodesic is indeed a geodesic. In addition, we have the following equivalent characterization of the geodesic in a geodesic space.

\begin{lem}
Let $(X, d)$ be a geodesic space, $p > 1$, and $\omega : [0,1] \to X$ a curve connecting $x_0$ and $x_1$. Then the followings are equivalent.
\begin{enumerate}
    \item $\omega$ is a constant-speed geodesic.
    
    \item $\omega \in \text{AC}(X)$ such that for almost every $t \in [0, 1]$, we have
    \begin{equation}
        |\omega'|(t) = d(\omega(0), \omega(1)).
    \end{equation}
    
    \item $\omega$ solves
    \begin{equation}
        \min \left\{ \int_0^1 |\tilde\omega'|^p \, \rd t : \tilde\omega(0) = x_0, \tilde\omega(1) = x_1 \right\}.
    \end{equation}
\end{enumerate}
\end{lem}

\subsection{Wasserstein space}

Since the Wasserstein space $({\cal P}_p(\bR^d), W_p)$ for $p \geq 1$ is a metric space, the following definition specializes Definition~\ref{def:metric_derivative} to the Wasserstein metric derivative.

\begin{defn}[Wasserstein metric derivative]
\label{def:wasserstein_metric_derivative}
Let $\{\mu_t\}_{t \in [0,1]}$ be an absolutely continuous curve in the Wasserstein (metric) space $({\cal P}_p(\bR^d), W_p)$. Then the \emph{metric derivative} at time $t$ of the curve $t \mapsto \mu_t$ with respect to $W_p$ is defined as
\begin{equation}
    |\mu'|_p(t) : = \lim_{h \to 0} {W_p(\mu_{t+h}, \mu_{t}) \over |h|}.
\end{equation}
For $p = 2$, we write $|\mu'|(t) := |\mu'|_2(t)$.
\end{defn}

In the rest of this section, we consider probability measures $\mu_t$ that are absolutely continuous with respect to the Lebesgue measure on $\bR^d$ and we use $\mu_t$ denote the probability measure, as well as its density, when the context is clear. 

\begin{thm}
\label{thm:metric_derivative_cty_eqn}
Let $p > 1$ and assume $\Omega \in \bR^d$ is compact.

\underline{\bf Part 1.} If $\{\mu_t\}_{t \in [0,1]}$ is an absolutely continuous curve in $W_p(\Omega)$, then for almost every $t \in [0,1]$, there is a velocity vector field $\vv_t \in L^p(\mu_t)$ such that
\begin{enumerate}
    \item $\mu_t$ is a weak solution of the continuity equation $\partial_t \mu_t + \div(\mu_t \vv_t) = 0$ in the sense of distributions (cf. the definition in~\eqref{eqn:cty_eqn_weak_solution} below);
    
    \item for almost every $t \in [0, 1]$, we have
    \begin{equation}
        \|\vv_t\|_{L^p(\mu_t)} \leq |\mu'|_p(t),
    \end{equation}
    where $\|\vv_t\|_{L^p(\mu_t)}^p = \int_{\Omega} \|\vv_t\|_2^p \, \rd \mu_t$.
\end{enumerate}

\underline{\bf Part 2.} Conversely, if $\{\mu_t\}_{t \in [0,1]}$ are probability measures in ${\cal P}_p(\Omega)$, and for each $t \in [0, 1]$ we suppose $\vv_t \in L^p(\mu_t)$ and $\int_0^1 \|\vv_t\|_{L^p(\mu)} \, \rd t < \infty$ such that $(\mu_t, \vv_t)$ solves the continuity equation, then we have
\begin{enumerate}
    \item $\{\mu_t\}_{t \in [0,1]}$ is an absolutely continuous curve in $({\cal P}_p(\bR^d), W_p)$;
    
    \item for almost every $t \in [0, 1]$,
    \begin{equation}
        |\mu'|_p(t) \leq \|\vv_t\|_{L^p(\mu_t)}.
    \end{equation}
\end{enumerate}
\end{thm}

As an immediate corollary, we have the following dynamical representation of the Wasserstein metric derivative.

\begin{cor}
\label{cor:metric_derivative_cty_eqn}
If $\{\mu_t\}_{t \in [0,1]}$ is an absolutely continuous curve in $({\cal P}_p(\bR^d), W_p)$, then the velocity vector field $\vv_t$ given in Part 1 of Theorem~\ref{thm:metric_derivative_cty_eqn} must satisfy
\begin{equation}
    \|\vv_t\|_{L^p(\mu_t)} = |\mu'|_p(t).
\end{equation}
\end{cor}

Corollary~\ref{cor:metric_derivative_cty_eqn} suggests that $\vv_t$ can be viewed as the \emph{tangent vector field} of the curve $\{\mu_t\}_{t \in [0,1]}$ at time point $t$. Moreover, Corollary~\ref{cor:metric_derivative_cty_eqn} suggests the following (Euclidean) gradient flow for tracking particles in $\bR^d$: let $y(t) := y_x(t)$ be the trajectory starting from $x \in \bR^d$ (i.e., $y(0) = x$) that evolves according the ordinary differential equation (ODE)
\begin{equation}
\label{eqn:particle_ode}
    {\rd \over \rd t} y(t) = \vv_t( y(t) ).
\end{equation}
The dynamical system~\eqref{eqn:particle_ode} defines a flow $Y_t : \Omega \to \Omega$ of vector field $\vv_t$ on $\Omega$ via
\begin{equation}
    \label{eqn:flow_representation}
    Y_t(x) = y(t).
\end{equation}
Then, it is straightforward to check that the pushforward measure flow $\mu_t := (Y_t)_\sharp \mu_0$ and the chosen velocity vector field $\vv_t$ in the ODE~\eqref{eqn:particle_ode} is a weak solution of the continuity equation $\partial_t \mu_t + \div(\mu_t \vv_t) = 0$ in the sense that
\begin{equation}
\label{eqn:cty_eqn_weak_solution}
    {\rd \over \rd t} \int_\Omega \phi \, \rd t = \int_\Omega \langle \nabla \phi, \vv_t \rangle \, \rd \mu_t,
\end{equation}
for any ${\cal C}^1$ function $\phi : \Omega \to \bR$ with compact support.

\begin{thm}[Constant-speed Wasserstein geodesic]
\label{thm:wasserstein_geodesic}
Let $\Omega \in \bR^d$ be a convex subset and $\mu, \nu \in {\cal P}_p(\Omega)$ for some $p > 1$. Let $\gamma$ be an optimal transport plan under the cost function $\|x-y\|_p^p$. Define
\begin{align*}
    & \pi_t : \Omega \times \Omega \to \Omega, \\
        & \pi_t(x, y) = (1-t) x + t y,
\end{align*}
as the linear interpolation between $x$ and $y$ in $\Omega$. Then, the curve $\mu_t = (\pi_t)_\sharp \gamma$ is a constant-speed geodesic in $({\cal P}_p(\bR^d), W_p)$ connecting $\mu_0 = \mu$ and $\mu_1 = \nu$.
\end{thm}

If $\mu$ has a density with respect to the Lebesgue measure on $\bR^d$, then there is an optimal transport map $T$ from $\mu$ to $\nu$~\citep{Brenier1991}. According to Theorem~\ref{thm:wasserstein_geodesic}, we obtain \emph{McCann's interpolation}~\citep{MCCANN1997153} in the Wasserstein space as
\begin{equation}
    \mu_t = [(1-t) \text{id} + t T]_\sharp \mu,
\end{equation}
which is a constant-speed geodesic in $({\cal P}_p(\bR^d), W_p)$. $\text{id}$ is the identity function in $\bR^d$.

To sum up, we collect the following facts about the geodesic structure and dynamical formulation of the OT problem. Let $p > 1$, and $\Omega \subset \bR^d$ be a convex subset (either compact or have no mass escaping at infinity).

\begin{enumerate}
    \item The metric space $({\cal P}_p(\Omega), W_p)$ is a geodesic space.
    
    \item For $\mu, \nu \in {\cal P}_p(\Omega)$, a constant-speed geodesic $\{\mu_t\}_{t \in [0, 1]}$ in $({\cal P}_p(\Omega), W_p)$ between $\mu$ and $\nu$ (i.e., $\mu_0 = \mu$ and $\mu_1 = \nu$) must satisfy $\mu_t \in \text{AC}({\cal P}_p(\Omega))$ and
    \begin{equation}
        |\mu'|(t) = W_p(\mu(0), \mu(1)) = W_p(\mu, \nu)
    \end{equation}
    for almost every $t \in [0, 1]$.
    
    \item The above $\mu_t$ solves 
    \begin{equation}
        \min \left\{ \int_0^1 |\tilde\mu'|^p(t) \, \rd t : \tilde\mu(0) = \mu, \tilde\mu(1) = \nu, \tilde\mu \in \text{AC}({\cal P}_p(\Omega)) \right\}.
    \end{equation}
    
    \item The above $\mu_t$ solves the Benamou-Brenier problem
    \begin{equation}
        W_p^p(\mu, \nu) = \min \left\{ \int_0^1 \|\vv_t\|_{L^p(\tilde\mu_t)}^p \, \rd t : \tilde\mu(0) = \mu, \tilde\mu(1) = \nu, \partial_t \tilde\mu_t + \div(\tilde\mu_t \vv_t) = 0 \right\},
    \end{equation}
    and $\mu_t = \mu(\cdot, t)$ defines a constant-speed geodesic in $({\cal P}_p(\Omega), W_p)$.
\end{enumerate}

\section{Entropic regularization}
\label{app:sec:Entropic_regularization}

 Our GeONet is compatible with entropic regularization, which is closely related to the Schr\"odinger bridge problem and stochastic control~\citep{ChenGeorgiouPavon2016}. Specifically, the entropic-regularized GeONet (ER-GeONet) solves the following fluid dynamic problem:
 
 \vspace{-5mm}
\begin{align}
\label{eqn:benamou-brenier_entropic_regularization}
\begin{gathered}
\min_{(\mu, \vv)} \int_0^1 \int_{\mathbb{R}^d} {1\over2} || \vv(x, t) ||_2^2 \ \mu(x, t) \ \rd x \ \rd t \\
  \qquad\qquad\qquad\qquad \mbox{subject to}  \ \  \partial_t \mu + \div(\mu \vv) + \varepsilon \Delta \mu = 0, \ \ \mu(\cdot, 0) = \mu_0, \ \ \mu(\cdot, 1) = \mu_1.
 \end{gathered}
\end{align}

Applying the same variational analysis as in the unregularized case $\varepsilon = 0$ (cf. Appendix~\ref{app:sec:primal-dual_optimality}), we obtain the KKT conditions for the optimization~\eqref{eqn:benamou-brenier_entropic_regularization} as the solution to the following system of PDEs:

\vspace{-5mm}
\begin{align}
    \label{eqn:benamou-brenier_entropic_regularization_kkt_1}
      \partial_t \mu + \div(\mu \nabla u) = & -\varepsilon \Delta \mu, \\
      \label{eqn:benamou-brenier_entropic_regularization_kkt_2}
      \partial_t u + {1\over2} \|\nabla u\|_2^2 = & \ \ \varepsilon \Delta u,
\end{align}
\vspace{-4mm}

with the boundary conditions $\mu(\cdot,0) = \mu_0, \mu(\cdot,1) = \mu_1$ for $\varepsilon > 0$. Note that~\eqref{eqn:benamou-brenier_entropic_regularization_kkt_2} is a parabolic PDE, which has a unique smooth solution $u^\varepsilon$. The term $\varepsilon \Delta u$ effectively regularizes the (dual) Hamilton-Jacobi equation in~\eqref{eqn:benamou-brenier_kkt}. In the zero-noise limit as $\varepsilon \downarrow 0$, the solution of the optimal entropic interpolating flow~\eqref{eqn:benamou-brenier_entropic_regularization} converges to solution of the Benamou-Brenier problem~\eqref{eqn:benamou-brenier_formula} in the sense of the method of vanishing viscosity~\citep{Mikami2004,evans2010}.

\section{Gradient enhancement}
\label{Method_augmentations}

 In practice, we may fortify the base method by adding extra residual terms of the differentiated PDEs to our loss function of GeONet. Such gradient enhancement technique has been used to strengthen  PINNs~\citep{YU2022114823}, which improves efficiency as fewer data points are needed to be sampled from $U(\Omega) \times U(0,1)$, and prediction accuracy as well.

The motivation behind gradient enhancement stems minimizing the residual of a differentiated PDE. We turn our attention to PDEs of the form
\begin{equation}
\label{eqn:gradient_enhancement_1}
\left\{
\begin{gathered}
\mathcal{F} \Big(x, t, \partial_{x_1}u, \hdots, \partial_{x_d}u, \partial_{x_1 x_1} u , \hdots, \partial_{x_d x_d} u, \hdots, \partial_t u, \lambda \Big) = 0 \ \ \ \ \text{on} \ \ \ \ \Omega \times [0,1], \\ \ \ \ \ \ \ \ \ \ \ \ \ \ \ \ \ \ \ \ \ \ \ \ \ \ \ \ \ \ \ \ \ \ \ \ \ 
u(\cdot, 0) = u_0, \ \ \ u(\cdot, 1) = u_1 \ \ \ \ \ \ \ \ \ \ \ \ \  \text{on} \ \ \ \ \Omega, \ \ 
\end{gathered}
\right.
\end{equation}
for domain $\Omega \subseteq \mathbb{R}^d$, parameter vector $\lambda$, and boundary conditions $u_0, u_1$. One may differentiate the PDE function $\mathcal{F}$ with respect to any spatial component to achieve
\begin{equation}
\label{gradient_enhancement_derivative}
\frac{ \partial }{ \partial x_{\ell}} \mathcal{F} \Big(x, t, \partial_{x_1}u, \hdots, \partial_{x_d}u, \partial_{x_1 x_1} u , \hdots, \partial_{x_d x_d} u, \hdots, \partial_t u, \lambda \Big) = 0 . 
\end{equation}
The differentiated PDE is additionally equal to $0$, similar to what we see in a PINN setup. If we substitute a neural network into the differentiated PDE of~\eqref{gradient_enhancement_derivative}, what remains is a new residual, just as we saw with the neural network substituted into the original PDE. Minimizing this new residual in the related loss function characterizes the gradient enhancement method. 

We utilize the same loss function in~\eqref{eqn:GeONet_loss}, but we add the additional terms
\begin{align}
\label{eqn:gradient_enhancement_2}
\mathcal{L}_{\text{GE,cty},i} \ \ \ & = \ \ \ \frac{1}{N} \sum_{\ell=1}^d \gamma_{\ell} | |  \ \ \frac{\partial}{\partial x_{\ell}} (\frac{\partial}{\partial t} \mathcal{C}_{\phi,i} + \div ( \mathcal{C}_{\phi,i} \nabla \mathcal{H}_{\psi,i}) )  \ \ | |_{L^2(\Omega \times (0,1))}^2,  \\
\mathcal{L}_{\text{GE, HJ},i} \ \ \ & = \ \ \  \frac{1}{N} \sum_{\ell=1}^d \omega_{\ell} | | \ \ \frac{\partial}{\partial x_{\ell}} ( \frac{\partial}{\partial t} \mathcal{H}_{\psi,i} + \frac{1}{2} | | \nabla \mathcal{H}_{\psi,i}| |_2^2 )  \ \  | |_{L^2( \Omega \times (0,1))}^2, 
\end{align}
where the variables and neural networks that also appeared in~\eqref{eqn:GeONet_loss} are the same. Here $\gamma_{\ell}$ and $\omega_{\ell}$ are positive weights. The summation is taken in order to account for gradient enhancement of each spatial component of $x \in \Omega$.

\begin{comment}
\begin{table}[h]
\caption{Error results are listed with gradient enhancement added, performed with 100 test pairings.}
  \label{tab:Gradient_Enhancement_CIFAR-10}
  \centering
  \begin{tabular}[b]{
    l
    S[table-format = 3]
    S[table-format = 2]
    S[table-format = 1.3]
    S[table-format = -2.2]
    S[table-format = 1.3]
    S[table-format = 1.3]
    S[table-format = 2.2]
    }
    \toprule
    \multicolumn{1}{c}{} & 
    \multicolumn{5}{c}{Gradient-enhanced CIFAR-10 results}\\
    \cmidrule(lr){2-6}        
    \textbf{CIFAR-10 data} & {$\bm{ t=0 }$} & {$\bm{  t=0.25  }$} & {$\bm{ t=0.5  }$} & {$\bm{ t=0.75  }$} & {$\bm{ t =1 }$}  \\
    \hline
    \rule{0pt}{2.5ex}  
    $\int_{\Omega} |\Upsilon_{\zeta^*} - u|^2 dx$ \ \ \ \ &
    {$0.39 \pm 0.33$}  & {$2.6 \pm 2.8$}  & {$4.4 \pm 5.1$} & {$2.9 \pm 2.9$} & {$0.57 \pm 0.61$}  \\
    \bottomrule
\end{tabular}
\end{table}
\end{comment}

\section{DeepONets}
\label{app:sec:DeepONets}

A challenge resides in solving the risk minimization problem over numerous instances of data. This challenge may be conciliated by instituting a DeepONet that learns a general nonlinear operator, where one (or a pair of) neural network(s) encode(s) the input and another encodes the collocation samples. This architecture originates as an equivalence to the universal approximation theorem for operators.

\textbf{General DeepONet.} A general operator $G^{\dagger}$ may be approximated by an unstacked DeepONet~\citep{392253,LuJinPangZhangKarniadakis2021_DeepONet}

\vspace{-2mm}
\begin{equation}
\label{eqn:DeepONet}
G^{\dagger}( u_0) (x,t) \approx \sum_{k=1}^{p} \mathcal{B}_{k} \big( u_0(x_1), \hdots, u_0(x_m), \theta \big) \cdot \mathcal{T}_{k} (x, t, \xi) , 
\end{equation}

where $\mathcal{B}_{k}, \mathcal{T}_{k}$ are scalar elements of output of neural networks $\mathcal{B}, \mathcal{T}$, and $p$ is a constant denoting the number of such elements. We take $\mathcal{B}$ and $\mathcal{T}$ to be artificial neural networks parameterized by $\theta, \xi$ respectively. $\mathcal{B}, \mathcal{T}$ are known as the branch and trunk networks respectively. $u_0$ is the initial function in which the operator is applied, evaluated at distinct locations $x_1, \hdots, x_m$ for branch input. $(x,t)$ is any arbitrary point in space and time in which $G^{\dagger}(u_0)$ may be evaluated.

\textbf{Enhanced DeepONet.} The above framework is restricted to one initial input function $u_0$. We turn our attention to the enhanced DeepONet, a DeepONet styled to act upon dual initial conditions~\cite{https://doi.org/10.48550/arxiv.2202.08942}. Our true operator $\Gamma^{\dagger}$ may be approximated using a second neural network encoder for input $u_1$,

\vspace{-2mm}
\begin{equation}
\label{eqn:DeepONet_enhanced}
\Gamma^{\dagger}( u_0, u_1) (x,t) \approx \sum_{k=1}^{p} \mathcal{B}^0_{k} \big( u_0(x_1), \hdots, u_0(x_m), \theta^0 \big) \cdot
\mathcal{B}^1_{k} \big( u_1(x_1), \hdots, u_1(x_m), \theta^1 \big) \cdot
\mathcal{T}_{k} (x, t, \xi) .
\end{equation}

\textbf{Physics-informed DeepONet.} The enhanced DeepONet may be substituted into any physics-informed framework, such as that of equation~\eqref{eqn:PINN_formulation_1}, taking place of the PDE solution value in the empirical loss to be minimized~\cite{DBLP:journals/corr/abs-2103-10974}.  Generalization of the trained DeepONet permits any solution to the PDEs to be evaluated instantaneously given the appropriate input function(s).

\section{Specialized aerchitectures}

\subsection{Modified multi-layer perceptron}
\label{Modified_mlp}

Here we outline the forward pass of the modified multi-layer perceptron used throughout the bivariate and MNIST encoder experiments. Let $\sigma$ denote an activation function (typically $\tanh$), $X$ as neural network design input, $W^i$ the weights of the neural network at index $i$, and $b^i$ the bias at index $i$. Here, $X$ can refer to either branch or trunk inputs, as this architecture is used for both.

The forward pass is given by
\begin{align}
& U = \sigma(W^1 X + b^1), \ \ \ V = \sigma(W^2 X + b^2) \\ 
& H^{1} = \sigma( W^{h,1} X + b^{h,1} ) \\ 
& Z^{k} = \sigma( W^{z,k} H^k + b^{z,k} ) \\ 
& H^{k} = (1 - Z^{k-1} ) \odot U + Z^{k-1} \odot U \\ &
\mathcal{N}_{\theta} = W^{\ell} H^{\ell} + b^{\ell} ,
\end{align}
where $k \in \{1,\hdots,\ell\}$, $\odot$ is an element-wise product, and $\mathcal{N}_{\theta}$ is the neural network final output, either a branch or a trunk.

\subsection{Fourier feature architecture}
\label{Fourier_feature}

We outline the Fourier feature architecture of our empirical Gaussians experiment of~\ref{empirical_Gaussians}. We leave the branches as multi-layer perceptrons, but we modify the trunks. In particular, we use
\begin{align}
U_x & = ( \cos(2 \pi B_x x), \sin(2 \pi B_x x))^T \\
U_t & = ( \cos(2 \pi B_t t), \sin(2 \pi B_t t))^T \\
H_x^1 & = \sigma( W_x^1 U_x + b_x^1 ) \\
H_t^1 & = \sigma( W_t^1 U_t + b_t^1 ) \\
H_v^k & = \sigma( W_v^k H_v^{k-1} + b_v^k ), \ \ \ v \in \{x,t\} \\
H^{\ell} & = H_x^{\ell} \odot H_t^{\ell} \\
\mathcal{N}_{\theta} & = W^{\ell} H^{\ell} + b^{\ell} ,
\end{align}
where $k \in \{2,\hdots,\ell\}$, $\odot$ is an element-wise product, and $\mathcal{N}_{\theta}$ is the neural network trunk output. $B_v \in \mathbb{R}^{m \times d}$ are matrices, with the elements sampled from a $\mathcal{N}(0,\sigma_v^2)$ distribution. We denote $d$ the dimension of space or time, and $m$ is an arbitrary dimension inherent to the matrix, generally taken to be at least 100.

\newpage 

\section{Hyperparameter settings and training details}
\label{Hyperparameter_settings}

We discuss training characteristics of GeONet based on the primary experiments. An unmodified Adam optimizer was chosen for all branch, trunk neural networks with a learning rate starting from $5\mathrm{e}{-4}$. All layers share the same width. We use $\tanh$ activation for all neural networks.  Coefficients $\alpha_1,\alpha_2,\beta_0,\beta_1$ were computed after examining errors. Coefficients were selected in the range $[0.05,20]$. Neural network depths refer to $\ell$ in each modified MLP. Training is done on a NVIDIA T4 GPU.

\begin{table*}[h]
  \caption{Architecture and training details in our Gaussian mixture experiments of Section~\ref{sec:experiments} and Appendix~\ref{Training and performance}.}
\begin{center}
\begin{tabular}[H]{
l
S[table-format = 3]
S[table-format = 2]
S[table-format = 1.3]
S[table-format = -2.2]
S[table-format = 1.3]
S[table-format = 1.3]
S[table-format = 2.2]
}
\toprule
\multicolumn{1}{c}{Hyperparameter} & 
\multicolumn{1}{c}{1D Gaussians} &
\multicolumn{1}{c}{2D Gaussians} \\
\toprule
\ \ No. of initial conditions $(\mu_0, \mu_1)$ & {20,000} &  {5,000}       \\
\ \ $m$ (branch input dimension) & {100} &  {576}       \\
\ \ Branch width & {150} &  {200}       \\
\ \ Branch depth  & {7} & {7}     \\
\ \ Trunk width  & {100} & {150}       \\
\ \ Trunk depth & {7} & {7}    \\
\ \ $p$ (dimension of  outputs) & {800} & {800}\\
\ \ Batch size & {2,000} & {2,000}  \\
\ \ Final training time & {$\sim 2$ hrs} & {$\sim 2$ hrs} \\
\ \ Final training loss & {$\sim 1.5\mathrm{e}{-4}$} & {$\sim 1.8\mathrm{e}{-5}$}  \\
\ \ $\alpha_1,\alpha_2, \beta_0, \beta_1$ & {$0.5,0.25,1,1$} & {$0.5, 0.25, 1,1$}  \\
\bottomrule
\end{tabular}
\end{center}
\end{table*}

\begin{table*}[h]
  \caption{Architecture and training details in our empirical Gaussians and encoded MNIST experiments of Section~\ref{sec:experiments} and Appendix~\ref{Training and performance}.}
\begin{center}
\begin{tabular}[H]{
l
S[table-format = 3]
S[table-format = 2]
S[table-format = 1.3]
S[table-format = -2.2]
S[table-format = 1.3]
S[table-format = 1.3]
S[table-format = 2.2]
}
\toprule
\multicolumn{1}{c}{Hyperparameter} & 
\multicolumn{1}{c}{Empirical Gaussians} & \multicolumn{1}{c}{Encoded MNIST} \\
\toprule
\ \ No. of initial conditions $(\mu_0, \mu_1)$  &  {1,000}  & {30,000}      \\
\ \ $m$ (branch input dimension) &  {625}  & {32}      \\
\ \ Branch width &  {100}  & {150}     \\
\ \ Branch depth   & {7} & {7}    \\
\ \ Trunk width   & {100}  & {100}      \\
\ \ Trunk depth  & {5} & {7}    \\
\ \ $p$ (dimension of  outputs)  & {200} & {200} \\
\ \ Batch size  & {1,000} & {1,000} \\
\ \ Final training time  & {$\sim 2$ hrs} & {$\sim 4$ hrs} \\
\ \ Final training loss  & {$\sim 7.0\mathrm{e}{-4}$} & {$\sim 2.0\mathrm{e}{-2}$} \\
\ \ $\alpha_1,\alpha_2, \beta_0, \beta_1$  & {$0.5, 0.25, 1,1$} & {$1,1,1,1$} \\
\bottomrule
\end{tabular}
\end{center}
\end{table*}

\newpage 
\section{Training and performance}
\label{Training and performance}

\subsection{Univariate and bivariate Gaussian mixture experiments}

\textbf{Univariate Gaussians.} We choose spatial domain $x \in \Omega = [0,10]$ discretized into a $100$ point mesh. We generate $20,000$ training pairs $(\mu_0, \mu_1)$ of Gaussians, taking $k_j = 6$ for the number of Gaussians in each mixture. We take means $\mu_i \in [2,8]$ and variances $\Sigma_i \in [0.5,0.6]$ uniformly. Empirically, we found a large batch size more suitable for training than a low one, and so we take a batch size of $2,000$, meaning this many uniform collocation points are taken for both the PDE residuals and boundary points for each training iteration. We choose physical loss coefficient $\alpha_1 = 0.5, \alpha_2 = 0.25$, with boundary coefficients $\beta_0 = \beta_1 = 1$. We found these coefficients a good balance to enforce the physical constraint without sacrificing boundary restrictions after iterating these coefficients among $[0.05,20]$ and examining error. Additional training details are given in Appendix ~\ref{Hyperparameter_settings}.

\begin{figure}[h]
  \centering
  \includegraphics[scale=0.56]{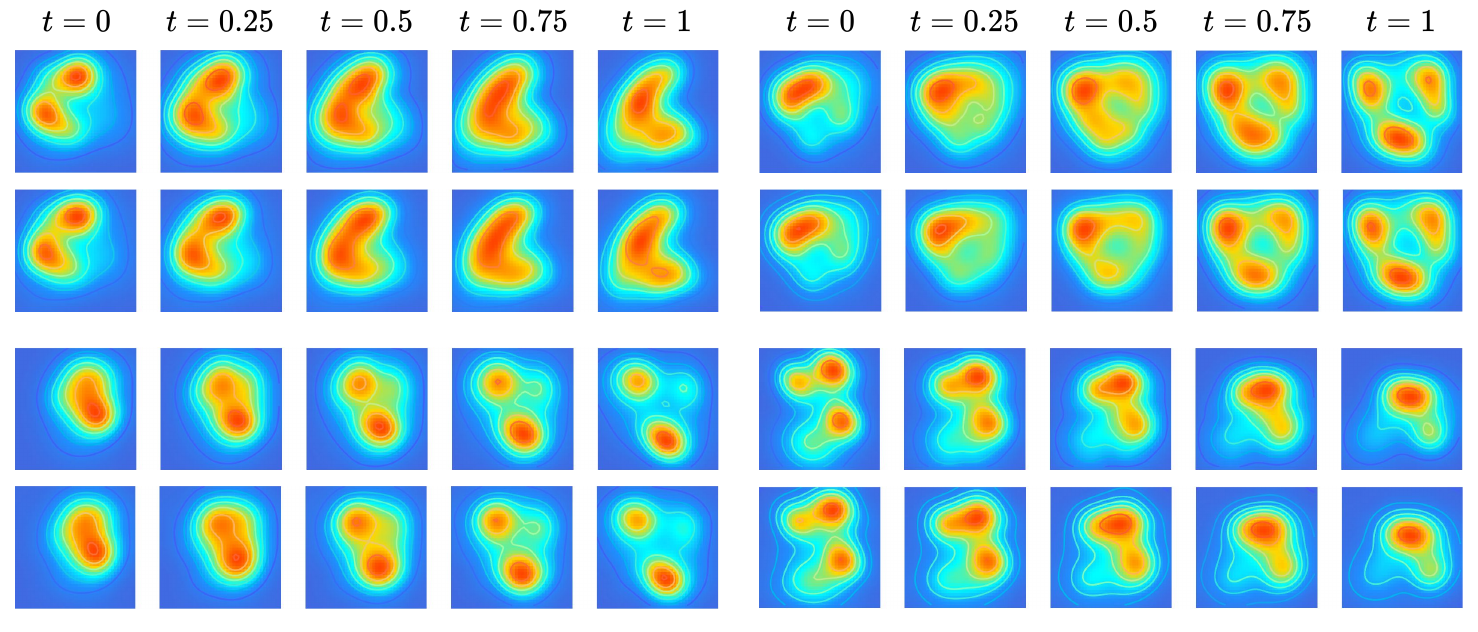}
  \caption{Geodesics predicted by GeONet on bivariate Gaussians over a square domain. The top of each pair is the reference solution computed by POT, and the bottom is GeONet.  \vspace{-3mm}}
  \label{fig:GeONet_gaussian_mixture_bivariate}
\end{figure}

\textbf{Bivariate Gaussians.} In our experiment, domain $\Omega = [0,5] \times [0,5] \subseteq \mathbb{R}^2$ was chosen, which was discretized into a $24 \times 24$ grid for GeONet input, meaning the branch networks took vector input of $576$ in length for each. We generate $5,000$ training pairs $(\mu_0, \mu_1)$. Recall that GeONet is mesh-invariant, so the $24 \times 24$ grids can be adapted to any higher resolution, which is used in figure~\ref{fig:GeONet_gaussian_mixture_bivariate}. We use a combination of low and high variance Gaussians in the mixture, 6 of which had variance in $[0.35,0.4]$ and 6 in $[0.75, 0.9]$, giving a total of 12 Gaussians in each mixture in each pair. Covariance was in $[-0.1,0.1]$.
Additional training details are given in Appendix ~\ref{Hyperparameter_settings}.

\textbf{Training.} To compute the DeepONet derivatives, we take the inner product in the enhanced DeepOnet as in equations~\eqref{eqn:Continuity_sol},~\eqref{eqn:HJ_sol}, and subsequently use automatic differentiation after the inner products are taken. Alternatively, one may compute a Hessian for the second-order derivatives, but this is costly in terms of memory, meaning a large batch size cannot be used without a monumental memory cost. We found the neural networks do not train properly without a large batch size, and so this method of differentiation is not viable. We found the DeepONet output dimension taken to be quite large  slightly outperforms a lower-dimension output given sufficient data and no overfitting. In the univariate Gaussian experiment, we take $p=800$, which outperformed $p=200$ by reducing training loss from approximately $2.5\mathrm{e}{-4}$ to $1.5\mathrm{e}{-4}$ and reducing test error by about $1\%$. In the bivariate experiment, changing $p=400$ to $p=800$ reduced training loss from approximately $2.1\mathrm{e}{-5}$ to $1.8\mathrm{e}{-5}$. Architecture generally made some difference to training loss, but not significant, making a width of around $100$-$200$ suitable for branches and trunks. For example, increasing branch width in the univariate experiment from $100$ to $150$ lowered training loss by approximately $4\mathrm{e}{-5}$. Increasing branch width to $200$ and trunk width to $150$ from $150$ and $100$ respectively had minimal effect, lowering training loss by about $1\mathrm{e}{-5}$. We found the modified MLP architecture preferable, lowering final training loss from approximately $3\mathrm{e}{-4}$ with standard architecture for univariate Gaussians.

\textbf{Performance.} Our baseline results were collected by deploying GeONet on the identity geodesic in Table~\ref{tab:GeONet_gaussian_mixture}. The baseline identity geodesic provides a benchmark for comparing and interpreting the errors across different setups. The univariate cases were evaluated upon a $100$ point mesh, and the bivariate upon a $40 \times 40$ mesh, except in the zero-shot super resolution case, in which the grid is refined and previously specified. From Table~\ref{tab:GeONet_gaussian_mixture}, we can draw the following observations. The loss boundary conditions~\eqref{eqn:GeONet_loss_BC} allow greater precision for $t=0,1$, which suggests that lack of data-enforced conditions along the inner region of the time continuum would cause greater error. Errors for predicting the univariate Gaussian trivial identity geodesic in the intermediate $t = 0.25, 0.5, 0.75$ are uniformly smaller than other in-distribution setups since the former is an easier task. In the bivariate experiment, we found that  error quickly rises as variance decreases, which is equivalent to a task of learning more complicated geodesics. We did not find lower variance drastically affects performance in the univariate experiment, suggesting GeONet and potentially physics-informed DeepONets in general are less effective as dimension increases. We did not find the number of Gaussians in the mixtures drastically affected results, but naturally more complicated geodesics induce greater error, which is to be expected. We found bivariate errors are similar to the random case as in the identity case, suggesting there is some notion of base neural operator error, which may not exist with simpler data.

\subsection{Gaussian empirical densities}

\textbf{Training.} 3000 point cloud particles were sampled from mixtures composed of 3 Gaussians for source $\mu_0$ and target $\mu_1$. 2D histograms were constructed to turn particle data into empirical densities, with bins ranging from $-7$ to $7$. Domain $\Omega = [0,5] \times [0,5]$ was discretized into a $25 \times 25$ point domain and assigned for the histograms' locations used as GeONet spatial input. A batch size of 1,000 was chosen. We take $p=200$, $\alpha_1 = 0.5, \alpha_2 = 0.25, \beta_0 = \beta_1 = 1$, which can be altered to impose strength of the boundary and physics terms accordingly. We employ the Fourier feature network architecture of~\cite{Wang_2021} for trunk networks. We take matrix $B_v$ with elements sampled in $\mathcal{N}(0,\sigma_v^2)$, subsequently taking $(\cos(2 \pi B_v v), \sin(2 \pi B_v v))^T$ as input for a fully-connected network, where $v$ is either space or time input. Our architecture for this experiment is fully outlined in ~\ref{Fourier_feature}. Empirically, we found low variance necessary, and we chose $\sigma = 0.5$ for both $v = x,t$ for both continuity and trunk branches.

\textbf{Performance.} In this experiment, GeONet is the only method among the comparison which correctly captures the geodesic behavior. While the other methods are suitable for point clouds specifically, they fail to capture the geodesic. This is demonstrated by our errors in table~\ref{tab:GeONet_gaussian_mixture_discrete}, as the errors of GeONet are significantly smaller than those of other methods. GeONet tends to slightly regularize the solutions, but the Fourier feature embedding does help correct this by being able to introduce more detail, such as the randomness from the samples, into the solution, such as with the initial conditions. Generally, we found higher variance of $B_v$ failed, as the physics-informed terms fail. For example, the Hamilton-Jacobi equation learns the trivial solution. High variance generally is able to capture detail to a better degree than low variance in the Fourier feature architecture ~\cite{tancik2020fourier}. A possible remedy for this is to train the networks on low variance, then fix the Hamilton-Jacobi equation once converged. Then, one may possibly retrain only the continuity networks.

\subsection{MNIST experiment}

\begin{table*}[!b]
\caption{$L^1$ error of GeONet on 50 test pairings of encoded MNIST. All values are multiplied by $10^{-2}$. Error was calculated upon the geodesic in both the shifted and ambient/original space.}
  \label{tab:GeONet_CIFAR-10}
  \centering
  \begin{tabular}[h]{
    l
    S[table-format = 3]
    S[table-format = 2]
    S[table-format = 1.3]
    S[table-format = -2.2]
    S[table-format = 1.3]
    S[table-format = 1.3]
    S[table-format = 2.2]
    }
    \toprule
    \multicolumn{1}{c}{} & 
    \multicolumn{5}{c}{GeONet $L^1$ error on encoded MNIST data}\\
    \cmidrule(lr){2-6}        
    \textbf{Test setting} & {$\bm{ t=0 }$} & {$\bm{  t=0.25  }$} & {$\bm{ t=0.5  }$} & {$\bm{ t=0.75  }$} & {$\bm{ t =1 }$}  \\
    \midrule
    \text{Encoded, identity} &
    {$0.923 \pm 0.213$} &
    {$0.830 \pm 0.166$} & 
    {$0.825 \pm 0.165$} & 
    {$0.834 \pm 0.173$} & 
    {$0.931 \pm 0.215$} \\
    \text{Encoded, random} & 
    {$1.62 \pm 0.333$} & 
    {$2.14 \pm 1.22$} & 
    {$2.78 \pm 1.62$} & 
    {$2.11 \pm 1.17$} & 
    {$1.54 \pm 0.282$} \\
    \midrule
    \text{Ambient, identity} & 
    {$26.7 \pm 11.2$} & 
    {$34.0 \pm 6.88$} & 
    {$35.3 \pm 8.32$} & 
    {$36.4 \pm 9.77$} & 
    {$34.0 \pm 13.2$} \\
    \text{Ambient, random} & 
    {$32.1 \pm 16.6$} & 
    {$58.2 \pm 15.0$} & 
    {$68.1 \pm 18.8$} & 
    {$56.4 \pm 14.3$} & 
    {$24.7 \pm 10.7$} \\
    \bottomrule
\end{tabular}
\vspace{-3mm}
\end{table*}

\textbf{Training.} To learn the geodesic, we ensure all values within the encoded representation are nonnegative, meaning we can shift all encoded representations by some arbitrary constant. We choose $10$ for this. This constant can be deducted in later stages to ensure the valid represenation is met. A domain of $[0,5]$ was dividied into an equispaced mesh of $32$ points for the encoded representation. This domain is rather arbitrary and is chosen simply for DeepONet input purposes, which can be modified as seen fit. $5,000$ encoded pairs were chosen to train GeONet, but the pretrained autoencoder was done on the entirety of MNIST with a batch size of $500$ and $120$ iterations per epoch. Additional training details are found in Appendix ~\ref{Hyperparameter_settings}.

\begin{figure}[h]
  \centering
  \vspace{-2mm}
  \includegraphics[scale=0.75]{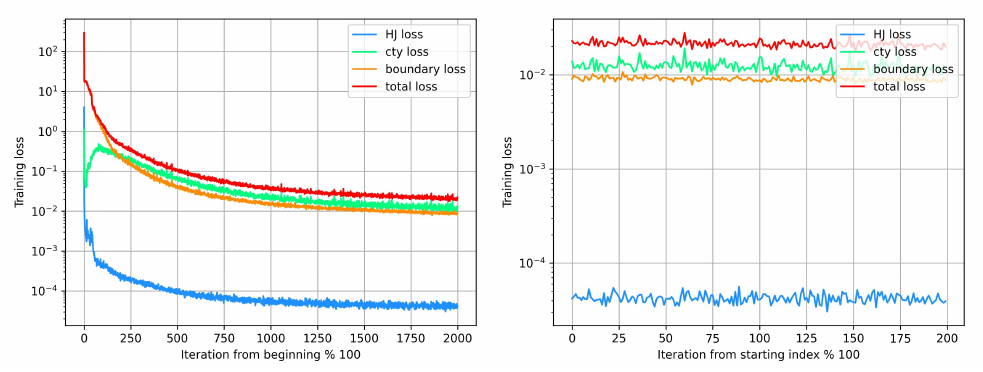}
  \caption{We examine iterations of the Adam optimizer in the total and late training on a log scale. The learning rate was modified every iteration. We examine late training in order to observe oscillatory behavior between the continuity and Hamilton-Jacobi loss to see if they adversarially compete in late training. We do not observe this pattern, and the continuity loss greatly surpasses the HJ loss in value. These graphs were created using the encoded MNIST experiment. \vspace{-3mm}}
  \label{fig:training_loss_late_training}
\end{figure}

\textbf{Performance.} GeONet performs well in this experiment. Scaling the physics-informed term by a constant less than one did not prove necessary in this experiment to ensure all loss terms are met to a sufficient degree. As before, boundary terms are uniformly smaller, likely since these terms are known and included in the loss function to be minimized. The same error metric is used as in the synthetic experiments but with normalization, making the $L^1$ error relative. We remark OOD generalization is omitted because the distribution of the encoded data is not known. We also remark the decoded images, being the geodesic returned to its original state, do not directly translate to a geodesic performed upon an original pair of images. NaN values are omitted in the error computations, which are possible in the POT solutions due to the irregularity of the initial conditions. 

\textbf{Regularization.} Classical geodesic algorithms require a small regularization parameter in order to be computed. This provides affects the synthetic experiments trivially, but we found this regularization induces greater in the MNIST experiment. This is to be considered when evaluating the errors, and true error is likely smaller between GeONet and the reference geodesics computed with POT than what is displayed. This regularization acts as a form of "smoothing" of the solutions.

\begin{comment}
\subsection{Out-of-distribution generalization.}
\begin{wrapfigure}{r}{0.45\textwidth}
	
\vspace{-8pt}
  \centering
    \includegraphics[width=0.42 \textwidth]{OOD_Univariate_1.pdf}
    \vspace{-1.1mm}
  \caption{An illustration of an OOD univariate Gaussian mixture geodesic at test time.}
  \label{fig:OOD_Univariate}
\end{wrapfigure}
%------------------------------------------

To examine the generalization performance of GeONet under distribution shifts, we consider a setup which is out-of-distribution (OOD) at test time. To do this, we alter the variances to be lower than in the training setup. Changing the number of Gaussians in the mixture at test time does not yield a significant enough of an alteration to the experiment. For univariate Gaussians, variance was among $[0.3,0.4]$ instead of the $[0.5,0.6]$ in training. For bivariate Gaussians, we take variances in $[0.25,0.3]$ and $[0.65,0.8]$, six of each in each mixture. Empirically, we found OOD generalization had lower error in this experiment than in the univariate case, which may be because the support of areas of low variance have relatively low measure. This bivariate error does indeed scale greater as variance decreases even more. We remark the lower variance generally yields a more sophisticated task in learning the geodesic, and error even trained upon such data is higher than with high variance.
\end{comment}

\section{GeONet error for additional error metrics}
\label{app:error_metrics}

\textbf{Error metric.} We use the $L^1$ error $\int_{\Omega} | \mathcal{C} - \mu | \rd x$ as our error metric to assess the performance, where $\mu := \mu(x, t)$ is a reference geodesic as proxy of the true geodesic without entropic regularization. The $L^1$ error integral is estimated by evaluating a discrete Riemann sum along a mesh and the reference is computed using the Convolutional Wasserstein Barycenter framework within the POT Python library~\citep{Solomon_2015,flamary2021pot}. Since $\int_{\Omega} | \mu| \rd x = 1$ for all time points, the $L^1$ error is also a relative error, a meaningful metric essentially corresponding to the percentage error between the neural operator geodesic and the reference. We also consider the $L^2$ and Wasserstein error metric for predicted Wasserstein geodesics. %Relative $L^2$ error is a common metric among comparisons, but we do not have previous geodesic results to compare, in which our proposed metric is more suitable.

\begin{table}[H]
  \caption{We list mean and standard deviations of error of GeONet on 50 random $\mu_0 \neq \mu_1$ samples for alternative error metrics, being $L^2$ error and the Wasserstein-1 distance. We remark we use sliced Wasserstein distance for the 2D case, as this metric is computationally feasible for higher dimensional cases. We perform this for random Gaussian mixture pairings. All values are multiplied by $10^{-2}$ by those of the table. \vspace{3mm}}
  % In the second part, we train and test upon $k_0 = k_1 = 5$, $\pi_i = 0.2$ for all $i$, with the same loss coefficients. 
  \label{tab:GeONet_gaussian_mixture_L2+W}
  \centering
  \begin{tabular}[b]{
    l
    S[table-format = 3]
    S[table-format = 2]
    S[table-format = 1.3]
    S[table-format = -2.2]
    S[table-format = 1.3]
    S[table-format = 1.3]
    S[table-format = 2.2]
    }
    \toprule
    \multicolumn{1}{c}{} & 
    \multicolumn{3}{c}{GeONet alternative metric error for random Gaussian mixtures}\\
    \cmidrule(lr){2-4}        
    \textbf{Experiment \ \ } & {$\bm{ t=0 }$} & {$\bm{  t=0.25  }$} & {$\bm{ t=0.5  }$}   \\
    \midrule
    \text{1D, $L^2$}  &
    {$5.19 \pm 1.74$}  & 
    {$6.91 \pm 4.81$}  & 
    {$7.28 \pm 5.39$} \\
    \text{1D, Wasserstein}  &
    {$0.352 \pm 0.116$}  & 
    {$0.364 \pm 0.178$}  & 
    {$0.403 \pm 0.228$} \\
    \midrule
    \text{2D, $L^2$}  &
    {$6.93 \pm 0.883$} & 
    {$7.72 \pm 1.23$} &
    {$8.11 \pm 1.30$} \\
    \text{2D, Wasserstein}  &
    {$0.245 \pm 0.0329$} & 
    {$0.264 \pm 0.0316$} &
    {$0.275 \pm 0.0447$}  \\

    \bottomrule
\end{tabular}

\vspace{3mm}

\begin{tabular}{
    p{2.5cm}  P{2.5cm} P{2.5cm} }
    \toprule
  %  \multicolumn{1}{c}{} 
  %  \multicolumn{5}{c}{GeONet alternative metric error for random Gaussian mixtures}\\
   % \cmidrule(lr){2-6}        
    \textbf{Experiment \ \ } &  {$\bm{ t=0.75  }$} & {$\bm{ t =1 }$}  \\
    \midrule
    \text{1D, $L^2$}  & 
    {$6.49 \pm 4.36$} & 
    {$4.81 \pm 1.58$} \\
    \text{1D, Wasserstein} & 
    {$0.386 \pm 0.166$} & 
    {$0.347 \pm 0.101$} \\
    \midrule
    \text{2D, $L^2$} & 
    {$7.79 \pm 1.14$} &
    {$6.87 \pm 1.05$} \\
    \text{2D, Wasserstein} & 
    {$0.267 \pm 0.0338$} &
    {$0.246 \pm 0.0356$} \\

    \bottomrule
\end{tabular}
%\vspace{-6mm}
\end{table}

\newpage
\section{Sample HJ graphs}

\begin{figure}[h]
  \centering
  \includegraphics[scale=0.55]{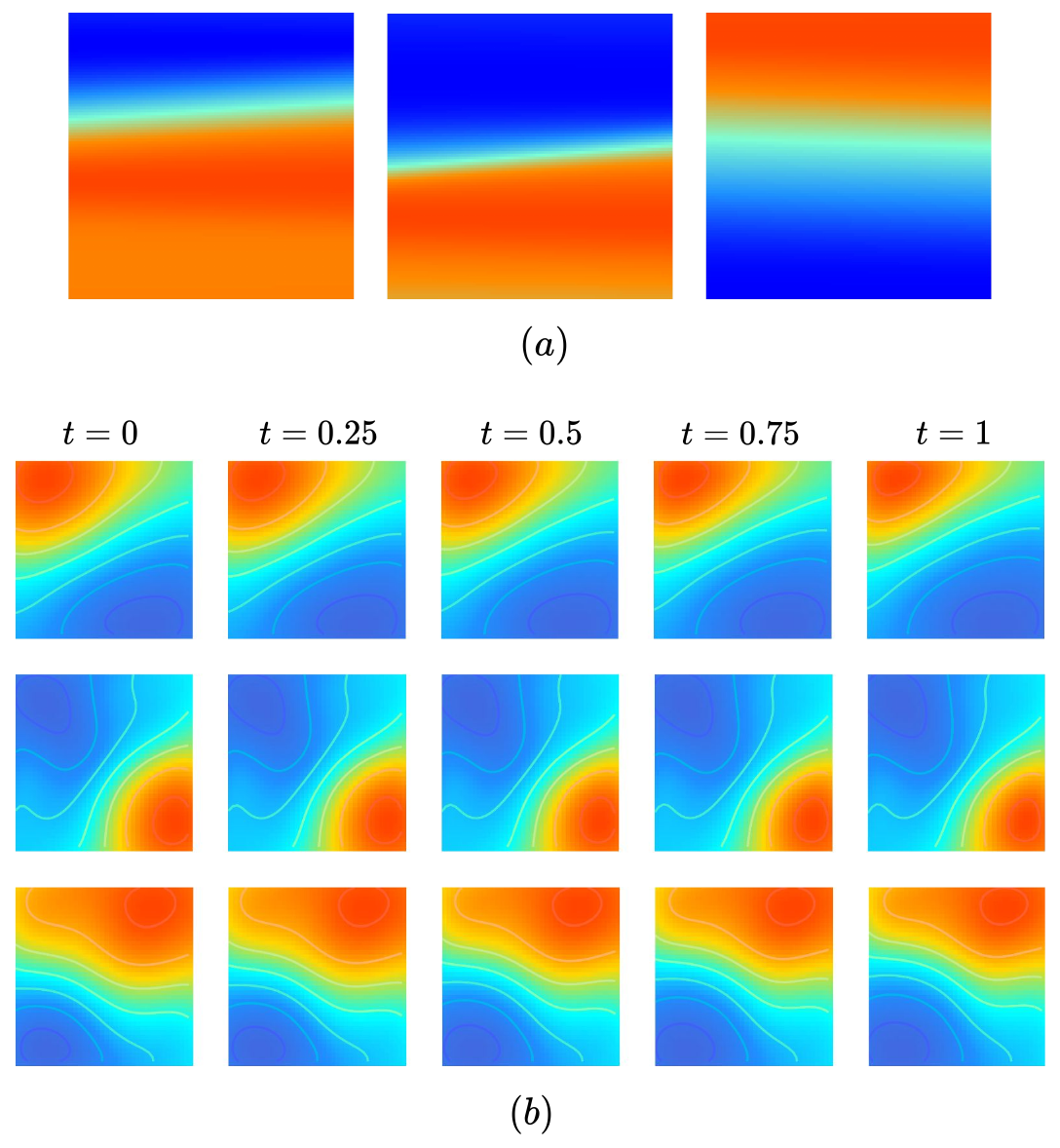}
  \caption{We present sample HJ equations for (a) three univariate Gaussian mixtures and (b) three bivariate Gaussian mixtures from the primary experiments performed in Section~\ref{sec:experiments}. The univariate HJ samples at certain times are the vertical cross sections of the graphs, and the bivariate samples are given at certain times. \vspace{-3mm}}
  \label{fig:GeONet_HJs}
\end{figure}

\end{document}